\documentclass[11pt]{article}

\usepackage{PRIMEarxiv}
\usepackage{tabularx}
\usepackage[utf8]{inputenc} %
\usepackage[T1]{fontenc}    %
\usepackage{url}            %
\usepackage{amsfonts}       %
\usepackage{nicefrac}       %
\usepackage{microtype}      %
\usepackage{lipsum}
\usepackage{fancyhdr}       %
\usepackage{graphicx}%
\usepackage[percent]{overpic}
\graphicspath{{media/}}     %
\usepackage{etoolbox}       %added by haoran
\usepackage{makecell}  % 导言部分引入makecell包
\usepackage{soul}
\usepackage{url}
\usepackage[hidelinks]{hyperref}
\usepackage[table]{xcolor}
\usepackage[utf8]{inputenc}
\usepackage{caption}
\usepackage{amsmath}
\usepackage{amssymb}
\usepackage{pifont}
\usepackage{wasysym}
\usepackage{booktabs}
\usepackage{algorithm}
\usepackage{algorithmic}
\usepackage[switch]{lineno}
\usepackage{listings}
\usepackage{xparse}
\usepackage{enumitem}
\usepackage{wrapfig}
\usepackage{adjustbox}
\usepackage{xspace}
\usepackage{listings}
\usepackage{subcaption}
\usepackage[most]{tcolorbox}

\usepackage{multicol}
\usepackage{multirow}
\usepackage{bbding}
\usepackage{circledsteps}

\newtcolorbox{promptbox}{
    colback=gray!5,          % 浅蓝色背景
    colframe=gray!70!black,  % 边框为深蓝色
    coltitle=white,          % 标题文字颜色为白色
    title=Case study - CoT Prompts example, % 标题
    fonttitle=\bfseries,     % 标题字体加粗
    colbacktitle=gray!70!black, % 标题栏为深蓝色
    boxrule=0.75pt,          % 边框厚度
    arc=3pt,                 % 圆角
    left=6pt,                % 左侧内边距
    right=6pt,               % 右侧内边距
    top=5pt,                 % 上侧内边距
    bottom=5pt,              % 下侧内边距
    boxsep=2pt,              % 内容和边框之间的距离
    fontupper=\sffamily\small  % 内容字体为无衬线小号字体
}

\usepackage{newtxtext, newtxmath}
\usepackage{threeparttable}%Table加reference脚注需要

\usepackage{xcolor}      % for \definecolor and \textcolor
\usepackage{tikz}        % for all the \tikzstyle, \tikzset, \node, \path, etc.
\usepackage{graphicx}    % for \scalebox
\usepackage{pifont}      % for \ding (used in your \cmark and \xmark)
\usepackage{xspace}      % for the trailing \xspace in your \ours macro
\usepackage{amsmath}

\AtBeginDocument{%
	\hypersetup{hidelinks}%
}
\usepackage{cleveref}
\definecolor{hidden-draw}{rgb}{0,0,0}
\definecolor{mycolor}{RGB}{248,233,204}
\definecolor{customblue}{RGB}{138,180,210}
\usepackage{adjustbox}
\usepackage{forest}
\tikzstyle{my-box}=[
rectangle,
draw=hidden-draw,
rounded corners,
text opacity=1,
minimum height=1.5em,
minimum width=5em,
inner sep=2pt,
align=center,
fill opacity=.5,
line width=0.8pt,
]
\tikzset{
	leaf/.style={
		my-box,
		minimum height=1.5em,
		fill=mycolor, % Change the RGB values to the desired values
		text=black,
		align=left,
		font=\footnotesize,
		inner xsep=2pt,
		inner ysep=4pt,
		line width=0.8pt
	}
}

\usepackage{fancyhdr}
\usetikzlibrary{mindmap,shadows}
\usepackage[numbers]{natbib}
\usepackage{datetime}
\usepackage{tocloft}
\usepackage{lipsum}

\usepackage{tikz}
\usepackage{pgfplots}
\usepackage{xcolor}
\usepackage{forest}
\usepackage{tablefootnote}
\usepackage{url}
\usepackage{subfig}

\usepackage{amsfonts,amssymb}
\usepackage{color, soul}
\usepackage{mathrsfs}
\usepackage{cleveref}
\usepackage{float}
\usepackage{wrapfig}

\usepackage{amsthm}
\usepackage{bm}
\usepackage{bbm}
\usepackage{algorithm}
\usepackage{algorithmic}
\usepackage{colortbl}
\usepackage{enumitem}
\usepackage{color}
\usepackage{balance}
\usepackage{stfloats}
\usepackage{makecell}

\definecolor{deepred}{HTML}{CC3333}
\usepackage{amsfonts,amssymb}
\usepackage{bbm}
\usepackage{listings}
\definecolor{myblue}{HTML}{3399CC}
\definecolor{myred}{HTML}{993333}
\definecolor{boxblue}{HTML}{248f6b}
\definecolor{lightpink}{RGB}{255, 230, 230} % #FFCCCC
\definecolor{darkred}{RGB}{185, 38, 74}      % #CC0033
\usepackage{mdframed}
\usepackage{tcolorbox}
\usepackage{enumitem}
\usepackage{fontawesome}

\pgfplotsset{compat=1.18}

\makeatletter
\providecommand\sf@counterlist{}
\makeatother

\title{Generative AI in Transportation Planning: A Survey}

\author{
	{\bfseries Longchao Da$^{1}$}\quad
	{\bfseries Tiejin Chen$^{1}$}\quad
	{\bfseries Zhuoheng Li$^{1}$}\quad
	{\bfseries Shreyas Bachiraju$^{1}$}\quad
	{\bfseries Huaiyuan Yao$^{1}$}\quad
	{\bfseries Li Li$^{6}$}\quad \\
	{\bfseries Yushun Dong$^{5}$}\quad
	{\bfseries Xiyang Hu$^{1}$}\quad
	{\bfseries Zhengzhong Tu$^{3}$}\quad
	{\bfseries Dongjie Wang$^{4}$}\quad
	{\bfseries Yue Zhao$^{6}$}\quad
	{\bfseries Ben Zhou$^{1}$}\quad\\
	{\bfseries Ram Pendyala$^{1}$}\quad 
	{\bfseries Benjamin Stabler$^{2}$}\quad
	{\bfseries Yezhou Yang$^{1}$}\quad
	{\bfseries Xuesong Zhou$^{1}$}\quad
	{\bfseries Hua Wei$^{1}$*}\quad
	\\ 
	{\bfseries $^{1}$Arizona State University},
	{\bfseries $^{2}$DKS Associates},
	{\bfseries $^{3}$Texas A\&M  University},
	{\bfseries $^{4}$ University of Kansas}, \\
	{\bfseries $^{5}$ Florida State University}, 
	{\bfseries $^{6}$ University of Southern California}
	\vspace{-1pt}\\
}

\begin{document}
\maketitle
\renewcommand{\thefootnote}{\fnsymbol{footnote}}
% \footnotetext[1]{Corresponding author: Hua Wei}

% \footnotetext[1]{Latest Update: Mar., 2025.}
% \footnotetext[1]{Corresponding Author hua.wei@asu.edu}

\vspace{-2em}

\begin{abstract}
The integration of generative artificial intelligence (GenAI) into transportation planning has the potential to revolutionize tasks such as demand forecasting, infrastructure design, policy evaluation, and traffic simulation. However, there is a critical need for a systematic framework to guide the adoption of GenAI in this interdisciplinary domain. In this survey, we, a multidisciplinary team of researchers spanning computer science and transportation engineering, present the \textit{first} comprehensive framework for leveraging GenAI in transportation planning. Specifically, we introduce a new taxonomy that categorizes existing applications and methodologies into two perspectives: transportation planning tasks and computational techniques. From the transportation planning perspective, we examine the role of GenAI in automating descriptive, predictive, generative simulation, and explainable tasks to enhance mobility systems. From the computational perspective, we detail advancements in data preparation, domain-specific fine-tuning, and inference strategies such as retrieval-augmented generation and zero-shot learning tailored to transportation applications. Additionally, we address critical challenges, including data scarcity, explainability, bias mitigation, and the development of domain-specific evaluation frameworks that align with transportation goals like sustainability, equity, and system efficiency. This survey aims to bridge the gap between traditional transportation planning methodologies and modern AI techniques, fostering collaboration and innovation. By addressing these challenges and opportunities, we seek to inspire future research that ensures ethical, equitable, and impactful use of generative AI in transportation planning.
\end{abstract}

\newpage
\tableofcontents

\begin{tcolorbox}[colback=lightpink, colframe=darkred, title=\textbf{Important Notice}]
\begin{itemize}[leftmargin=*]

\item \textbf{Content Warning:} This paper may contain offensive content generated by LLMs. 

\end{itemize}
\end{tcolorbox}

\newpage

\begin{wrapfigure}[15]{r}{0.63\textwidth} 
	\centering
	\vspace{-1mm}
	\includegraphics[width=0.62\textwidth]{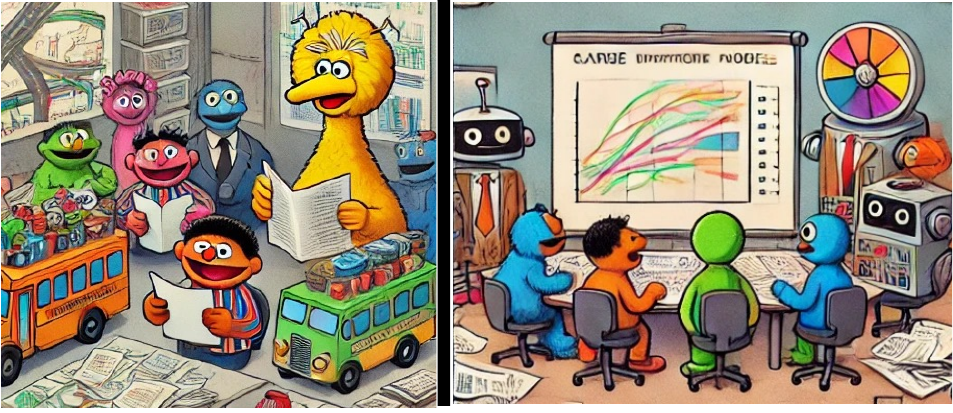}
	\caption{Generative AI is revolutionizing transportation planning through advanced language analysis and interdisciplinary integration capabilities. }
	\label{fig:cartoon}
\end{wrapfigure}

\section{Introduction}\label{sec:intro}
Recent advancements in generative artificial intelligence (GenAI) have showcased transformative potential across a wide array of fields, including healthcare~\cite{wornow2023shaky,kim2024health}, finance~\cite{huang2023finbert,wu2023bloomberggpt}, scientific discovery~\cite{zhang2023artificial,lu2024ai}, transportation~\cite{da2024open,zhang2024urban,zhang2024large}, and education~\cite{kasneci2023chatgpt}. Transportation planning, in particular, stands to benefit greatly from these advancements, as GenAI, particularly large language models (LLMs), offers tools to address complex challenges such as scenario generation~\cite{chang2024llmscenario}, multimodal system optimization~\cite{le2024multimodal}, stakeholder engagement~\cite{yu2024synthetic} and assistance with data analysis. By synthesizing insights from heterogeneous and dynamic data sources, LLMs have become essential tools for advancing transportation research and practice~\cite{zhao2023survey}.

Transportation planning is a systematic process of developing strategies to manage and enhance the movement of people and goods across various transportation systems while addressing long-term societal goals. This process integrates data-driven methodologies to balance efficiency, equity, and sustainability in mobility systems, accounting for multimodal networks, infrastructure needs, and policy constraints~\cite{bruton2021introduction,banister2008sustainable}. Transportation planning encompasses activities such as demand forecasting~\cite{banerjee2020passenger}, infrastructure design~\cite{parkin2012network}, traffic
management\cite{kurzhanskiy2015traffic}, and public engagement~\cite{de2014public}. Traditionally, transportation planning relied on expert-driven frameworks, where decision-makers analyzed travel patterns, forecasted demands, and designed solutions based on statistical models and simulation techniques~\cite{meyer2001urban}. However, these methods often struggle to manage the increasing scale and complexity of modern transportation systems, particularly in incorporating diverse data sources, addressing real-time dynamics, and generating adaptive solutions. They also struggle with the costs of technology-forward solutions with public agency budgets and the ability to pay salaries to keep technically strong talent.

\textbf{Motivating Example.} For instance, GenAI has revolutionized the generation of travel demand by synthesizing data from land use patterns, traffic counts, and environmental metrics to predict future infrastructure demands under varying conditions~\cite{wu2020spatiotemporal,chatterjee2023generating}.
Applications such as traffic simulation~\cite{zhang2024urban,zhong2023language} and policy sentiment modeling~\cite{tori2024performing} demonstrate GenAI's ability to enhance speed, accuracy, and scope, helping planners make informed decisions with confidence.

\textbf{Research Gaps.} Despite its potential, leveraging generative AI in transportation planning remains challenging due to three critical gaps, which require targeted research efforts:

\begin{itemize}[leftmargin=*] \item \textit{Lack of Systematic Integration Frameworks:} Existing studies offer limited guidance on systematically incorporating generative AI into transportation workflows, particularly for tasks such as multimodal travel optimization, real-time traffic management, or alternative scenario generation~\cite{ziems2024can}. While individual applications exist, there is no unified methodological framework to integrate AI-driven insights with established transportation models, making adoption inconsistent across different planning domains.
	
	\item \textit{Need for Transportation-Specific Model Adaptations:} General-purpose generative AI models struggle with domain-specific challenges, including data biases~\cite{he2023inducing}, hallucinated or unrealistic outputs~\cite{yao2023llm}, and high computational costs~\cite{marino2024integrating}. Transportation applications require tailored approaches such as scenario-specific fine-tuning, multimodal data fusion techniques, and computationally efficient architectures to ensure practical deployment in large-scale planning environments.
	
	\item \textit{Insufficient Domain-Specific Knowledge Integration:} General LLMs lack an intrinsic understanding of key transportation concepts, such as infrastructure constraints, multimodal travel behavior, and regulatory frameworks, limiting their effectiveness in real-world planning scenarios~\cite{mou2024unifying}. Addressing this gap requires specialized datasets, enhanced model training with transportation-focused priors, and hybrid AI approaches that integrate transportation simulation models with generative outputs. \end{itemize}

\textbf{Intended Audience.} This survey is intended for computer scientists, transportation researchers, interdisciplinary scholars, practitioners, and policymakers seeking to leverage generative AI for transportation planning.

\begin{figure*}[t!]
	\centering
	\begin{adjustbox}{width=0.90\textwidth}
		\begin{forest}
			for tree={
				grow=east,
				reversed=true,
				anchor=base west,
				parent anchor=east,
				child anchor=west,
				base=center,
				font=\large,
				rectangle,
				draw=hidden-draw,
				rounded corners,
				align=left,
				text centered,
				minimum width=5em,
				edge+={darkgray, line width=1pt},
				s sep=3pt,
				inner xsep=2pt,
				inner ysep=3pt,
				line width=0.8pt,
				ver/.style={rotate=90, child anchor=north, parent anchor=south, anchor=center},
			},
			where level=1{text width=16em,font=\normalsize,}{},
			where level=2{text width=22em,font=\normalsize,}{},
			where level=3{text width=13em,font=\normalsize,}{},
			where level=4{text width=7em,font=\normalsize,}{},
			[
			The Generative AI in \\ Transportation Planning, 
			[
			{Introduction (Sec. \ref{sec:intro})},
			],
			[
			Background (Sec. \ref{preliminary})
			[
			Transportation Planning (Sec. \ref{preliminary:1})
			],
			[
			{GenAI Methodologies (Sec. \ref{genaiMethods})}
			],
			[
			{Evolution of AI in Transportation (Sec. \ref{evolution-ai-transportation})}
			] 
			],
			[
			Classical Planning Functions and \\ Modern Transformations (Sec.~\ref{Traditional-Method})
			[
			Descriptive Tasks (Sec. \ref{descriptionTask})
			],
			[
			Predictive Tasks (Sec. \ref{predictiveTask})
			],
			[
			Generative Tasks (Sec. \ref{generativeTask})
			],
			[
			Simulation Tasks (Sec. \ref{simulationTask})
			],
			[
			Trustworthiness (Sec. \ref{trustworthy})
			]
			],
			[
			Technical Foundations for GenAI \\ in Transportation Planning (Sec.~\ref{Computer-Science})
			[
			Dataset Preparation (Sec. \ref{Benchmark})
			],
			[
			Fine-tuning GenAI Models  (Sec. \ref{fine-tuning})
			],
			[
			Zero-shot Inference with GenAI (Sec. \ref{zero-shot})
			],
			[
			Few-Shot Inference with GenAI (Sec. \ref{few-shot})
			],
			[
			Other Techniques Enhancing Inference (Sec. \ref{other-inference})
			],
			[
			Case Study: LLM for OD Calibration  (Sec. \ref{caseStudy}) 
			]
			],
			[
			Future Directions and \\ Challenges (Sec.~\ref{sec:challenges})
			[
			Pitfalls of GenAI for \\ Transportation Planning  (Sec. \ref{pitfalls})
			],
			[
			Pipelines Integrate Transportation \\ with GenAI  (Sec. \ref{pipelines})
			],
			[
			Data Scarcity and Domain-Specific \\ Datasets Construction (Sec. \ref{dataScarcity})
			],
			[
			Explainability and Hallucination (Sec. \ref{explain})
			],
			[
			Democratizing Access to \\ Transportation Knowledge  (Sec. \ref{democratizing})
			],
			[
			Call for Novel Evaluation Criteria  (Sec. \ref{evaluation})
			]
			],
			[
			Conclusion (Sec. \ref{sec:conclusion})
			]
			]
		\end{forest}
	\end{adjustbox}
	\vspace{-4mm}
	\caption{Organization of this survey.}
	\label{fig:tree}
\end{figure*}
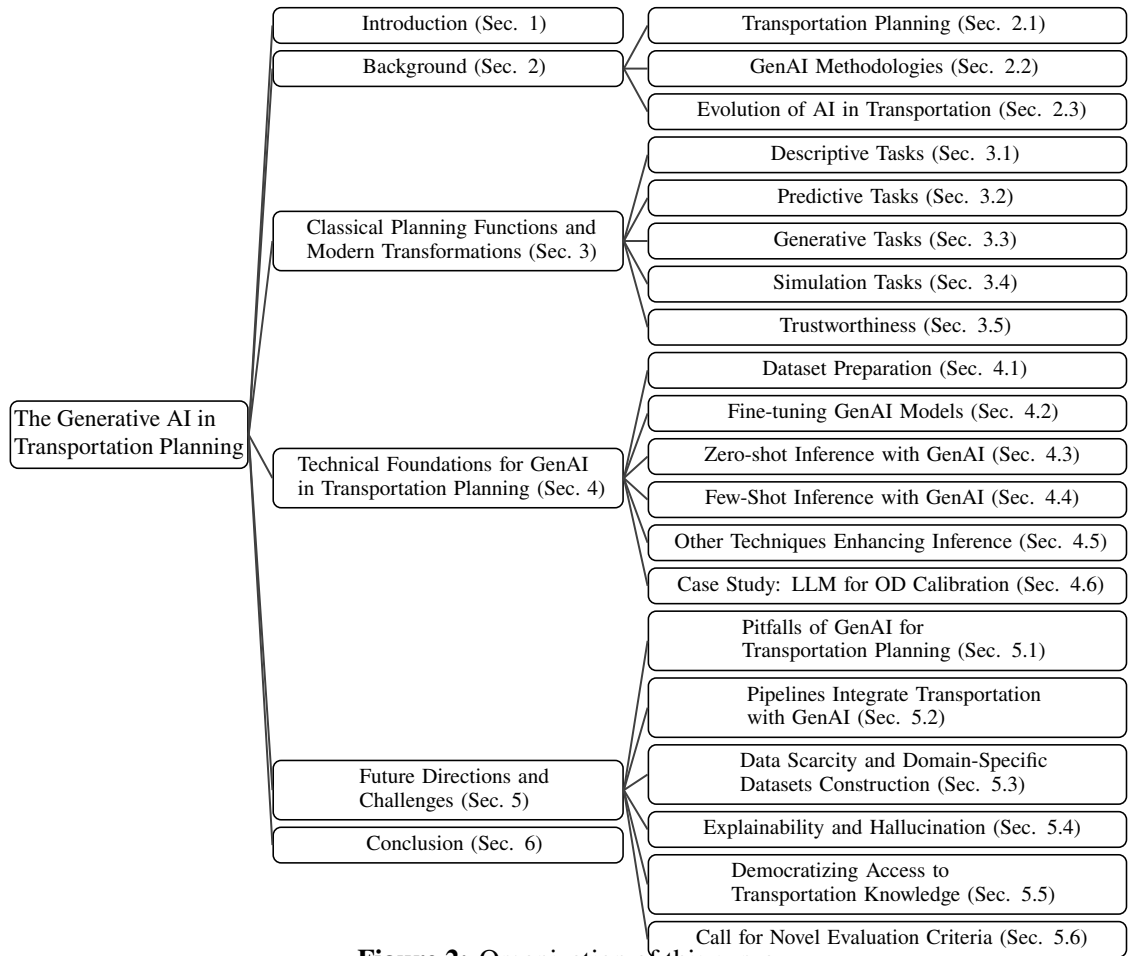

\textbf{Structure.} As shown in Figure~\ref{fig:tree}, Section~\ref{preliminary} provides background knowledge. Section~\ref{Traditional-Method} present classical transportation planning tasks and computational perspectives from GenAI, respectively. Section~\ref{Computer-Science} shows some application-specific applications to dataset and experimental results. Section~\ref{sec:challenges} discusses challenges and future directions, and Section~\ref{sec:conclusion} concludes the survey.

\section{Background} \label{preliminary}

\subsection{Transportation Planning}\label{preliminary:1}
Transportation planning is a structured process that supports the efficient movement of people and goods while addressing societal, environmental, and economic goals~\cite{bruton2021introduction,banister2008sustainable}. The process provides a roadmap for developing short-term and long-term strategies to optimize transportation systems, improve accessibility, and promote sustainability. Transportation planning includes a diverse set of activities, such as forecasting demand, evaluating infrastructure needs, optimizing traffic networks, and engaging stakeholders in decision-making processes.

\noindent \textbf{The 3C Planning Process.} Transportation planning follows the \textit{Continuing, Cooperative, and Comprehensive} (3C) framework: 
(1) \textbf{Continuing:} The planning process is ongoing and iterative, adapting to changing demographics, technology advancements, and travel behaviors. 
(2) \textbf{Cooperative:} Collaboration among local, regional, state, and federal agencies, along with public and private stakeholders, ensures shared goals and priorities. 
(3) \textbf{Comprehensive:} The process considers all transportation modes (e.g., road, rail, transit, pedestrian) and evaluates their impact on environmental, economic, and community systems~\cite{mihyeon2005addressing}. 
The 3C process provides a robust framework for ensuring transportation solutions are effective, adaptable, and inclusive.

\noindent \textbf{Performance-Based Approach.} Modern transportation planning emphasizes performance-based decision-making~\cite{neumann2004performance,pelorosso2020modeling,otte2021reference}, which links investments to measurable goals and outcomes. Performance metrics include travel time reliability~\cite{taylor2013travel}, safety improvements~\cite{chen2022network}, accessibility~\cite{abrishami2023comparing}, emissions reduction~\cite{buonocore2019metrics}, and equity~\cite{van2021evaluating}. By tracking these metrics, transportation planners ensure that infrastructure projects align with strategic goals, such as improving mobility, enhancing sustainability, and addressing community needs.

\noindent \textbf{Computational Methods.} Transportation planning employs a range of computational tools, data analytics, and predictive models to design and optimize systems for mobility and infrastructure. While traditional methods, such as static regression analysis, have historically played a central role in travel demand forecasting~\cite{yang2015development}, more advanced methodologies, like Activity-Based Models (ABMs), have significantly enhanced the scientific rigor and precision of demand modeling~\cite{national2015activity}. ABMs offer a disaggregated approach by simulating individual travel behaviors and activities, providing a more comprehensive and dynamic perspective on transportation systems. However, despite their sophistication, ABMs face challenges in handling the complexity and multimodal interactions of modern transportation systems~\cite{makarova2023role}.
Generative AI and other advanced computational approaches can augment ABMs by addressing these complexities. For example, generative AI can process large-scale heterogeneous data, simulate dynamic travel patterns, and support adaptive solutions that enhance the scalability and applicability of ABMs. By integrating these cutting-edge techniques, transportation planners can overcome existing limitations and enable broader adoption of ABMs in real-world contexts~\cite{vlahogianni2014short,cui2018deep,xu2024genai}.

\noindent \textbf{Key Tasks in Transportation Planning.}
Transportation planning is a multifaceted process that addresses mobility challenges while balancing goals such as efficiency, sustainability, equity, and economic development. Its scope encompasses a wide range of tasks:

\begin{figure}[b!]
	\centering
	\includegraphics[width=0.85\linewidth]{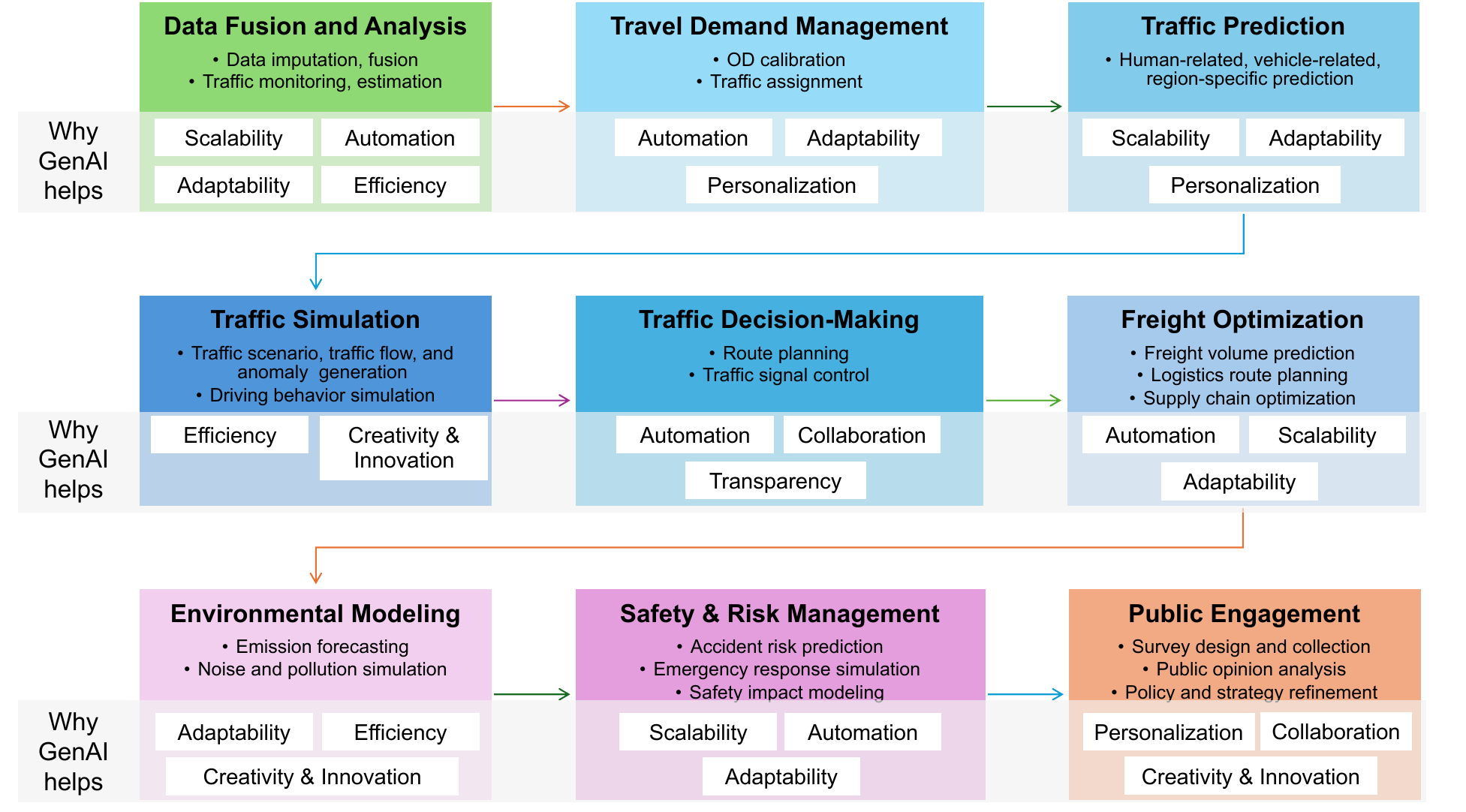}
	\caption{Tasks in transportation planning and potentials of Generative AI for these tasks. A \textit{partial} list of reasons on why GenAI helps transportation planning is also included.}
	\label{fig-why-genai}
\end{figure}

\begin{itemize}[leftmargin=*]
	\item \textit{Demand Forecasting:} Estimating travel demand for various transportation modes under future scenarios, accounting for human-related, vehicle-specific, and regional patterns~\cite{banerjee2020passenger,anguita2023air,ling2023sthan,zachariah2023systematic}.
	\item \textit{Data Fusion and Analysis:} Integrating diverse data sources, including traffic sensors, GPS trajectories, and public surveys, to enhance traffic monitoring, estimation, and decision-making~\cite{ounoughi2023data,zou2025deep}.
	\item \textit{Traffic Prediction:} Forecasting short- and long-term trends in human and vehicle mobility, with applications in congestion management and urban planning~\cite{zheng2020gman,liu2024spatial}.
	\item \textit{Traffic Simulation:} Modeling scenarios, traffic flows, and anomalies to analyze system performance under different operational and environmental conditions~\cite{zhang2024urban,zhong2023language, mei2024libsignal, da2024cityflower, da2023sim2real}.
	\item \textit{Traffic Decision-Making:} Optimizing route planning, traffic signal control, and multimodal system coordination to improve network performance~\cite{nha2012comparative,wei2018intellilight, da2023uncertainty, da2022crowdgail}.
	\item \textit{Environmental Modeling:} Simulating emissions, noise, and other environmental impacts to evaluate the sustainability of transportation strategies~\cite{ulengin2010problem}.
	\item \textit{Freight Optimization:} Enhancing logistics through freight volume prediction, route optimization, and supply chain coordination~\cite{archetti2022optimization}.
	\item \textit{Safety and Risk Management:} Predicting accident risks, simulating emergency responses, and modeling the safety impacts of transportation policies and designs~\cite{chen2022network,lin2019optimal, da2024shaded}.
	\item \textit{Public Engagement:} Designing and collecting public surveys, analyzing public opinion, and refining policies to align with stakeholder priorities~\cite{de2014public}.
	\item \textit{Performance Monitoring:} Tracking key performance indicators (KPIs) to evaluate the effectiveness and adaptability of transportation systems over time~\cite{garcia2018integrating,djordjevic2016key}.
\end{itemize}

This comprehensive task set emphasizes the critical role of data-driven methods and innovative tools in addressing the evolving challenges of transportation planning. The integration of generative AI further enhances these tasks by enabling scalability, adaptability, and automation in complex transportation scenarios. Figure~\ref{fig-why-genai} below presents a comprehensive framework for the emerging applications of Generative AI in transportation, including how Generative AI can possibly contribute to various stages of transportation planning and management. Each section of the diagram represents a crucial aspect of transportation operations, from data fusion and traffic management to public engagement and safety risk management. Generative AI can help transform these applications in several ways: 
\begin{itemize}[leftmargin=*]
	\item \textit{Scalability} enables AI models to process and analyze vast datasets, such as traffic monitoring, travel demand modeling, and infrastructure planning, across extensive geographic regions. This allows transportation planners to generalize insights across multiple locations while preserving spatial and temporal dependencies.
	\item \textit{Automation} streamlines time-consuming and labor-intensive tasks, such as route optimization, traffic signal control, and safety risk assessment, reducing the need for human intervention and improving operational efficiency. Unlike scalability, which focuses on handling large datasets, automation emphasizes minimizing manual effort in decision-making and system management. 
	
	\item \textit{Adaptability} ensures that generative AI can respond dynamically to changing traffic conditions, policy shifts, or behavioral patterns in real-time. Unlike automation, which focuses on predefined rule-based tasks, adaptability refers to the model’s ability to refine predictions or recommendations as new information becomes available, such as optimizing congestion mitigation strategies in response to weather disruptions.
	
	\item \textit{Efficiency} enhances decision-making speed and resource allocation by processing complex transportation data in real time. While automation reduces human workload, efficiency refers to AI’s ability to extract meaningful insights with optimized computational resources, reducing the time required to generate actionable recommendations.
	
	\item \textit{Personalization} tailors transportation services to individual needs by generating context-aware recommendations for different user groups, including freight logistics, public transit users, and active travelers. Unlike adaptability, which refers to system-level changes, personalization focuses on individual-level optimizations, such as providing route recommendations based on user preferences or accessibility requirements.
	
	\item \textit{Transparency and Explainability} improve stakeholder trust by generating interpretable insights in decision-making processes. While AI-generated outputs are not always fully verifiable due to potential biases and inconsistencies, efforts such as explainable AI techniques, causal reasoning, and retrieval-augmented generation (RAG) can enhance interpretability. However, it is important to acknowledge ongoing debates regarding the difficulty of verifying AI reasoning paths, as LLMs can generate plausible but misleading explanations.
	
	\item \textit{Creativity and Innovation} enable generative AI to propose novel transportation strategies by synthesizing diverse perspectives learned from extensive datasets. Unlike traditional optimization models, which focus on predefined criteria, AI can explore unconventional yet effective solutions, such as integrating autonomous shuttles into transit systems or designing resilient infrastructure for extreme weather events.
	
\end{itemize}

Overall, generative AI’s ability to bring together diverse data sources, enhance decision-making, and automate complex tasks positions it as a transformative tool in the future of transportation planning, helping agencies to improve efficiency, reduce risks, and meet the diverse needs of modern transportation systems.

\subsection{Generative AI Methodologies}\label{genaiMethods}

\begin{figure}[t!]
	\centering
	\vspace{-5mm}
	\includegraphics[width=0.9\textwidth]{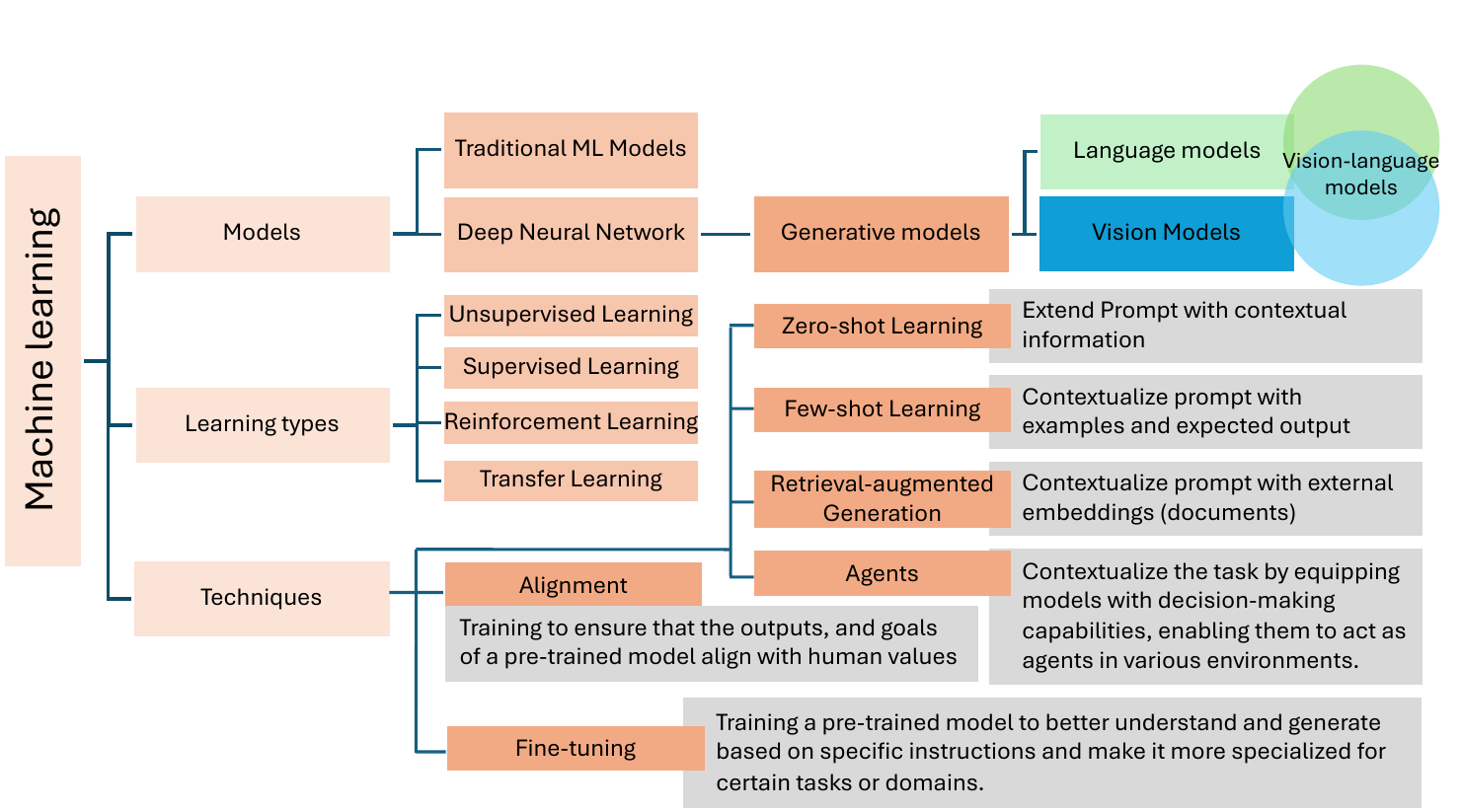}
	\caption{Generative AI techniques. The taxonomy is spread with three main categories, Models, Learning types, and Techniques, and each of the categories is followed with distinct aspects for comprehensive research domain coverage.}
	\label{fig:genai-tech}
\end{figure}

The methodologies underpinning generative AI applications in transportation include data preparation, fine-tuning, and inference techniques, which ensure that models are reliable, scalable, and tailored to domain-specific challenges.

\subsubsection{Generative AI Models}\label{genai-in-trans}

Generative AI models have demonstrated significant potential in advancing transportation planning and management. These models, including GANs (Generative Adversarial Networks), VAEs (Variational Autoencoders), diffusion models, LLMs (Large Language Models), and MLLMs (Multimodal Large Language Models), enable sophisticated data synthesis, simulation, and decision-making capabilities tailored to the complexities of transportation systems.

\noindent \textbf{GANs and VAEs.} GANs and VAEs are widely used for generating synthetic transportation data, such as traffic flows or multi-modal traffic network scenarios. GANs are particularly effective in creating high-quality, realistic datasets by training two networks—the generator and discriminator—against each other~\cite{goodfellow2014generative,arjovsky2017wasserstein}. For instance, GANs can simulate traffic patterns under rare conditions, such as extreme weather events, enabling planners to evaluate system resilience~\cite{cheng2023highway}. VAEs, on the other hand, excel in encoding transportation data into latent variables, providing interpretable representations that support efficient scenario generation and simulation~\cite{kingma2013auto}.

\noindent \textbf{Diffusion Models.} Recent advancements in diffusion models~\cite{ho2020denoising} have extended their application to transportation, offering robust capabilities for generating complex scenarios, such as adaptive routing plans or multimodal demand forecasts~\cite{peng2024diffusion}. These models iteratively refine noisy data inputs to generate realistic and contextually relevant outputs, making them ideal for tasks that require high-fidelity and granular predictions.

\noindent \textbf{Large Language Models (LLMs).} LLMs~\cite{achiam2023gpt,touvron2023llama,chiang2023vicuna}, such as GPT-based models, have revolutionized text-based transportation applications, including policy analysis, stakeholder engagement, and public sentiment analysis. Trained on vast corpora of text data, LLMs can interpret and generate natural language descriptions of transportation challenges, recommend actionable strategies, and facilitate decision-making~\cite{da2025survey}. For example, LLMs can assist in summarizing complex infrastructure plans for non-expert stakeholders, enhancing public understanding and participation. LLMs are also useful for data analysis assistance - like having a young smart data scientist helping analyze a data set. They need guidance, but they can create great charts and maps and tables and insights with human inputs.

\noindent \textbf{Multimodal Large Language Models (MLLMs).} MLLMs extend the capabilities of LLMs by incorporating additional data modalities, such as images, videos, and geospatial data. This multimodal integration allows MLLMs to handle complex transportation tasks, such as analyzing real-time traffic camera feeds or integrating sensor data with textual policy documents for comprehensive planning~\cite{liu2024visual,zhu2023minigpt,liu2024improved}. For instance, MLLMs can evaluate the visual conditions of roadways captured in images while generating textual recommendations for maintenance priorities or routing adjustments, bridging the gap between visual analysis and textual insights.
These models are further enhanced through advanced methods like fine-tuning, in-context learning, and retrieval-augmented generation to address domain-specific challenges in transportation planning. These models and methods go beyond traditional methods by capturing complex interdependencies within transportation data, offering significant improvements in areas such as demand forecasting, network optimization, and public sentiment analysis.

\subsubsection{Core Transportation Applications}\label{coreapp}

Generative AI supports various tasks in transportation planning, enhancing traditional methods with capabilities like scenario generation, demand forecasting, and traffic simulation.  

\textbf{Scenario Generation} leverages generative AI to explore alternative transportation strategies, such as infrastructure designs~\cite{xu2024leveraging, rane2023integrating, rane2023role, guridi2024image, onatayo2024generative}, policy interventions~\cite{kumar2024generative, balsa264leveraging, kumar2024generative, uddinrevolutionizing, paramesha2024enhancing}, and response plans for disruptive events~\cite{grigorev2024incidentresponsegpt, finkenstadt2024contingency, mondal2023bell}. For example, AI models can simulate the impacts of congestion pricing~\cite{ying2024beyond}, optimize transit-oriented developments~\cite{visvizi2024smart}, or model infrastructure resilience under extreme weather conditions~\cite{aasi2024generating, da2024shaded, camps2024ai, thulke2024climategpt}. By analyzing such scenarios, planners can identify cost-effective and environmentally sustainable solutions.  

\textbf{Demand Forecasting} applies generative AI to predict travel demand patterns across multimodal systems. Models can fine-tune origin-destination (OD) matrices to align with observed traffic counts~\cite{zhang2024trafficgpt,da2024open}, estimate the adoption of shared mobility services~\cite{yu2024large,li2024limp,ullah2024role}, and simulate long-term demand shifts due to demographic or economic changes~\cite{guo2024towards,liu2024spatial,movahedi2024crossroads,wang2024llm,tang2024large}. These forecasts provide insights into system bottlenecks, enabling planners to balance travel loads across transportation networks effectively.  

\textbf{Traffic Simulation and Optimization} focuses on modeling traffic dynamics and optimizing system performance. Generative AI enables simulations of mixed-autonomy systems, where human-driven and autonomous vehicles coexist, to improve vehicle coordination and reduce stop-and-go waves~\cite{yao2024comal}. Additionally, AI can optimize traffic signal timings~\cite{da2024prompt, wang2024llm} and route selection strategies~\cite{tran2023determining, tupayachi2024towards} to minimize delays and enhance urban mobility efficiency.  

\textbf{Sustainability and Resilience Planning} highlights the role of generative AI in advancing low-carbon and climate-resilient transportation systems. AI models can simulate eco-driving behaviors~\cite{xu2024genai, zhang2024car}, predict the adoption of electric vehicles~\cite{zhang2024role}, and evaluate infrastructure resilience under extreme scenarios, such as natural disasters~\cite{moreno2024generative, gedik5preliminary}. Further, generative tools can assess the accessibility of transportation systems to ensure equitable mobility solutions for underserved populations~\cite{sanguinetti2023using, yu2024large, ullah2024role}.

\textbf{Benchmark Datasets} form the foundation for evaluating generative AI performance in tasks like OD calibration, traffic simulation, and infrastructure planning. Examples include LargeST~\cite{liu2023largest} for traffic analysis and traffic assignment datasets~\cite{xu2024unified}. These datasets provide high-resolution inputs and ground-truth benchmarks to validate AI-driven insights. However, building comprehensive knowledge bases for these models remains a significant challenge. Particularly for forecasting tasks, where uncertainty and dynamic variables play key roles, such as: population growth, land-use changes, and evolving travel behaviors. Generative AI requires diverse, harmonized data inputs, such as historical traffic data, multimodal interactions, and prior model outcomes, but integrating these into a machine-readable format is time-intensive and complex. The lack of standardization, interoperability, and real-time updates further complicates the process, creating a bottleneck in AI adoption. Future advancements must focus on automated dataset synthesis, standardized modeling frameworks, and adaptive real-time updates to ensure that generative AI can address the intricate and evolving demands of transportation systems effectively.

\textbf{Data Preprocessing Strategies} ensure clean, balanced, and representative inputs for AI models. Transportation datasets often require manual or automated annotation of traffic patterns~\cite{moosavi2017annotation}, infrastructure attributes~\cite{wu2024efficient}, or user feedback~\cite{zhou2024openannotate2}. Generative AI methods, such as LLM-based automated labeling~\cite{ming2024autolabel, artemova2024hands}, address annotation challenges while preserving data quality. Data augmentation techniques, such as synthetic OD matrix generation~\cite{zhong2023estimating, pamula2023estimation}, further expand datasets to simulate rare or extreme transportation scenarios.

\textbf{Fine-Tuning Techniques} enable generative AI to adapt to transportation-specific tasks efficiently. Methods like Low-Rank Adaptation (LoRA)~\cite{wu2024dlora} optimize model performance while minimizing computational demands. For complex tasks like OD calibration, Chain-of-Thought prompting~\cite{yao2024comal} enhances the model's reasoning ability to align OD matrices with observed traffic counts.

\textbf{Generative Inference Techniques} allow AI models to address new tasks with limited training data. In zero-shot or few-shot learning, task-specific instructions or examples are embedded within prompts to guide the model’s outputs~\cite{li2024whattell}. Agent-based reasoning~\cite{guan2024richelieu} further enables generative AI to simulate dynamic interactions, such as vehicle coordination in mixed-autonomy systems or traffic merging behaviors under varying conditions.

\textbf{Advanced Inference Strategies} improve generative AI's adaptability to real-world transportation challenges. Retrieval-Augmented Generation (RAG) dynamically integrates external data sources, such as live sensor feeds or policy records, to produce accurate and context-aware predictions~\cite{wang2024evaluating, xu2024genai, dai2024vistarag, da2024evidencechat}. By modeling the interplay between agents and their environment, Graph-RAG~\cite{microsoft2024graphrag} has proven effective in multi-modal tasks like route planning~\cite{zhu2024challenges}, which can possibly be generalized to congestion management and safety-critical decision-making.
Besides, Self-Consistency Decoding reduces variability in generated outputs, stabilizing recommendations for tasks like traffic signal optimization or route planning~\cite{huang2023enhancing}.

\subsection{Evolution of AI in Transportation}
\label{evolution-ai-transportation}

The evolution of artificial intelligence (AI) in transportation reflects a transformative shift from manual and traditional approaches to AI-assisted and generative methodologies. This progression has redefined data processing, modeling, decision-making, and system validation, enabling transportation planning to become increasingly adaptive, data-driven, and automated.

\subsubsection{Traditional Approaches} Historically, traditional transportation planning heavily relied on expert-driven processes. Domain experts played a central role, manually collecting data through surveys and observations, constructing static models based on simplified assumptions, and validating plans using mental frameworks and iterative testing~\cite{meyer2001urban}. While effective for their time, these methods were resource-intensive and constrained by the limited ability to handle complex or dynamic systems.

\subsubsection{AI-assisted Approaches} The introduction of AI-assisted methodologies marked a significant leap forward. Machine learning, such as predictive models, enabled more efficient analysis of structured datasets, such as traffic counts, weather data, and household travel diary surveys and transit onboard surveys. AI-assisted systems provided predictions and optimization suggestions for tasks like congestion forecasting~\cite{akhtar2021review}, signal optimization~\cite{wei2018intellilight,wei2019colight}, and traffic rerouting~\cite{cao2016unified}. However, these systems required substantial human intervention for retraining, parameter tuning, and validation, making them adaptive but still reliant on expert oversight.

\subsubsection{GenAI Approaches} The latest evolution, generative AI—represents a paradigm shift toward highly autonomous systems. Generative models leverage large-scale, fine-grained datasets, including real-time sensor inputs and external sources like social media or weather forecasts~\cite{tsai2022traffic,guo2024towards}. These models autonomously generate solutions, simulate traffic scenarios, and optimize infrastructure designs without requiring explicit programming for each task~\cite{zhang2024urban,du2024large}. However, the integration of existing expert knowledge remains crucial in guiding these AI systems. Domain expertise is essential for defining objectives, validating outputs, and ensuring alignment with established transportation modeling principles. Expert insights help refine AI-driven processes by providing a clear direction, especially in complex tasks like multimodal demand forecasting or infrastructure planning, where understanding system nuances is critical. Generative AI systems, when combined with expert knowledge, enable rapid execution and real-time decision-making, accelerating tasks such as traffic rerouting during emergencies, infrastructure design under various constraints, and automated reporting for public engagement.

\begin{figure}[htbp] 
	\centering 
	\includegraphics[width=0.98\linewidth]{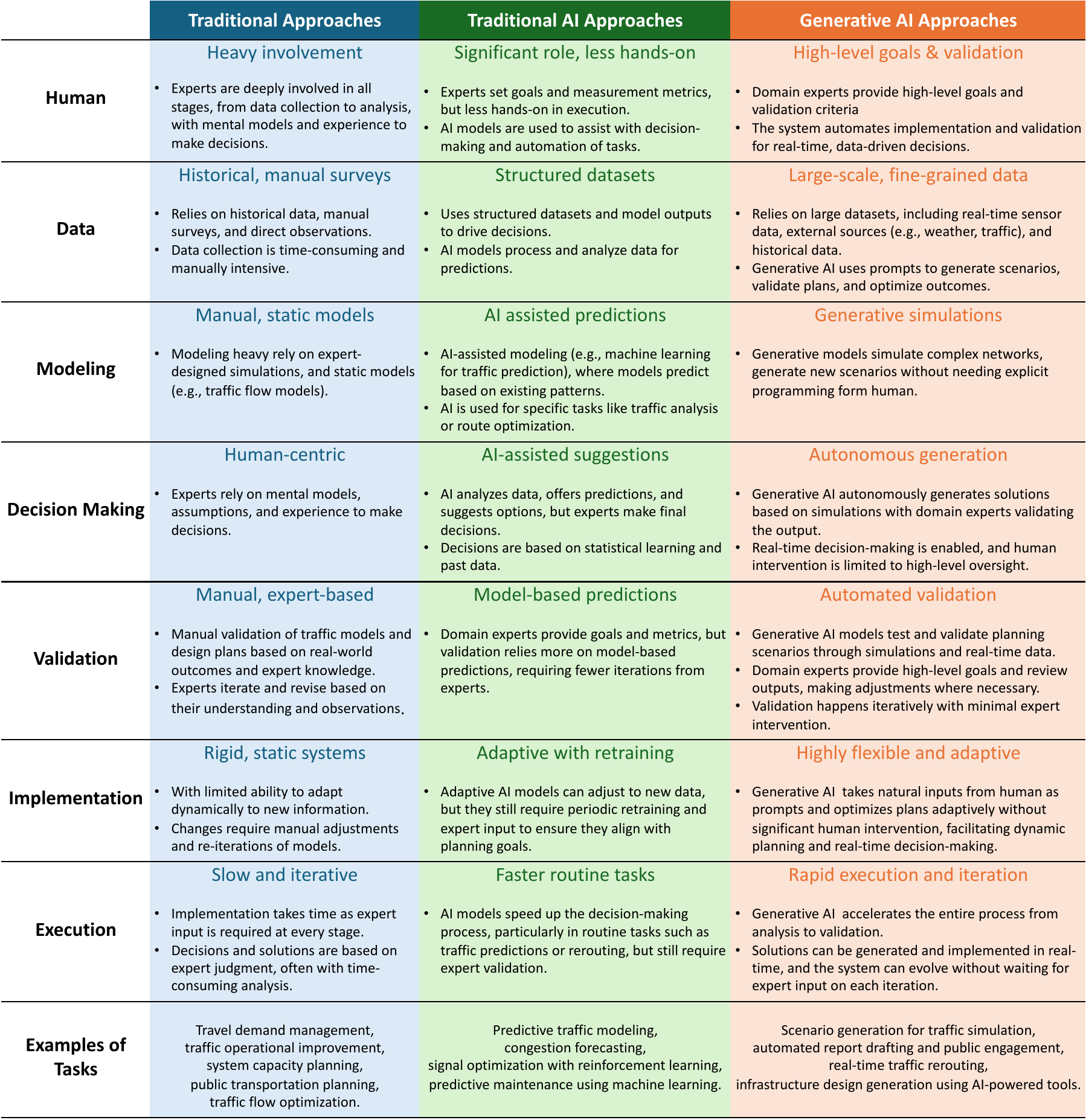} 
	\caption{Evolution of AI in Transportation: A Comparison of Traditional, AI-Assisted, and Generative Approaches.} 
	\label{fig-evolution-planning} \end{figure}

Figure~\ref{fig-evolution-planning} illustrates this evolutionary trajectory, comparing traditional, AI-assisted, and generative approaches across dimensions such as data processing, modeling, decision-making, validation, and implementation. For example, while traditional systems relied on historical data and manual validation, generative AI enables dynamic scenario generation and validation, empowering planners to respond adaptively to evolving conditions.

Generative AI’s potential to revolutionize transportation planning lies in its ability to integrate diverse datasets, adapt to dynamic environments, and generate actionable insights with minimal delay~\cite{guo2024towards,da2023llm,peng2024lc}. 
As generative AI continues to evolve, its integration with emerging technologies such as Internet of Things (IoT), 5G networks, and autonomous systems will further expand its applications. This evolution underscores the potential of generative AI to address complex challenges in modern transportation planning, paving the way for smarter, more equitable, and sustainable mobility solutions.

\section{Classical Transportation Planning Functions and Modern Transformations} \label{Traditional-Method}

Generative AI has introduced transformative advancements in transportation planning, reshaping traditional methods and enabling data-driven solutions to address increasingly complex challenges. In this section, we analyze how generative AI enhances five major categories of transportation planning tasks: \textit{descriptive tasks}, \textit{predictive tasks}, \textit{generative tasks}, \textit{simulation tasks}, and \textit{explainability}. Finally, we address societal and ethical implications, emphasizing the importance of equitable and transparent AI integration.

\subsection{Descriptive Tasks for Data Fusion and Analytics}\label{descriptionTask}
\textbf{Definition.} Descriptive tasks involve collecting, processing, integrating, and analyzing transportation-related data to extract actionable insights~\cite{wang2024visionllm}. 
These tasks serve as the foundation for all subsequent analyses, as they aim to summarize and describe existing conditions, identify patterns, and detect anomalies within complex transportation systems.

Generative AI significantly enhances descriptive tasks by automating data fusion and analytics from diverse, heterogeneous sources, including:
\begin{itemize}[leftmargin=*]
	\item \textit{Sensor Networks:} Real-time traffic flow and congestion data collected from GPS, loop detectors, and camera feeds~\cite{wang2024generative, sun2024genai}.
	\item \textit{User-Generated Data:} Crowdsourced data from ride-hailing platforms, navigation apps, and public transit feedback~\cite{manda2025current}.
	\item \textit{Infrastructure Data:} Road network topology, traffic signal operations, and transit schedules~\cite{xu2024genai}.
	\item \textit{Socioeconomic Data:} Demographic information, land use, and travel behavior surveys~\cite{zhou2024large, unlu2023chatmap}.
\end{itemize}

\noindent \textbf{Data Processing and Fusion.} Generative AI enables the seamless integration of multi-modal data, addressing inconsistencies such as missing values, noise, or overlapping data formats. Models like Generative Adversarial Networks (GANs)~\cite{yoon2018gain}, Variational Autoencoders (VAEs) and diffusion models are particularly effective for imputing incomplete data or fusing datasets to create coherent representations of transportation systems~\cite{kingma2013auto,ho2020denoising}. For example, OpenStreetMap data can be combined with GPS trajectory datasets to reconstruct road networks in under-mapped areas, enabling more accurate routing analysis~\cite{memduhouglu2024enriching, juhasz2023chatgpt, feng2024citybench}.  

\noindent \textbf{Descriptive Analytics for Pattern Detection.} Generative AI further supports the detection of temporal and spatial patterns in transportation systems. For instance:
\begin{itemize}[leftmargin=*]
	\item Traffic congestion hotspots can be identified by analyzing historical flow data, highlighting bottlenecks for targeted interventions~\cite{berhanu2023examining}.
	\item Public transit inefficiencies, such as delays or under-utilized routes, can be detected by analyzing GTFS data combined with passenger feedback~\cite{ devunuri2024transitgpt}.
	\item Urban mobility patterns, such as modal shifts during peak hours, can be extracted from GPS-based datasets to support multi-modal system design~\cite{chen2024deep,wang2021reinforced}.
\end{itemize}

These descriptive analytics not only summarize current conditions but also provide planners with baseline information for predictive and simulation-based tasks.

\noindent \textbf{Challenges.} Descriptive tasks face several challenges, including:
\begin{itemize}[leftmargin=*]
	\item \textit{Data Heterogeneity:} Integrating data from varied formats, resolutions, and sources remains a technical hurdle~\cite{zhang2024disttrain}. Transportation data often comes from diverse sources like GPS devices, traffic sensors, user apps, and public databases~\cite{oladimeji2023smart}.  And each of them has its unique structure and metadata. Standardizing these data streams into a unified format while preserving their integrity and context is essential but complex. 
	\item \textit{Scalability:} Processing large-scale datasets in real time requires efficient computational frameworks~\cite{xu2024large}. The increasing volume of transportation data, generated by millions of vehicles, sensors, and IoT devices, demands robust algorithms and infrastructure capable of handling such scale without compromising speed or accuracy~\cite{deekshith2023scalable}, and real-time even add on to its difficulty, where delays in data processing could lead to cascading inefficiencies.
	\item \textit{Privacy Concerns:} Handling user-generated or GPS data necessitates robust privacy-preserving mechanisms~\cite{wu2024new}. Sensitive information like travel patterns or location history, must be anonymized to protect user identities while still allowing meaningful analytics. How to balance between data utility and privacy compliance, especially under regulations like GDPR or CCPA~\cite{wong2023privacy}, requires innovative techniques such as differential privacy or secure multi-party computation
\end{itemize}

By addressing these challenges, generative AI can strengthen data analytics pipelines, ensuring that transportation planners have reliable and interpretable insights for informed decision-making.

\subsection{Predictive Tasks in Transportation Planning}\label{predictiveTask}
\textbf{Definition.} Predictive tasks involve forecasting transportation trends, such as traffic flow~\cite{wang2021exploring,pan2020spatio}, estimate time of arrival~\cite{wang2018will,liu2023uncertainty}, travel demand~\cite{wei2017integrating} or infrastructure performance~\cite{wei2018improving}, using historical and real-time data~\cite{tao2023towards}. Traditional methods, such as regression-based models~\cite{mahapatra2023regression} or rule-based simulations~\cite{tang2019cityflow}, often struggle to capture the dynamic and multi-dimensional nature of transportation systems. Generative AI overcomes these limitations by automating data preprocessing, enhancing prediction accuracy, and improving scalability~\cite{zhang2021traffic, da2024cityflower}.

Generative AI significantly accelerates predictive analysis by automating labor-intensive processes such as \textit{data annotation} and \textit{demand forecasting}. For instance, AI models can preprocess sensor data, detect anomalies, and predict congestion levels under varying conditions~\cite{liu2024poliprompt}. Tools such as fine-tuned LLMs have been applied to forecast multimodal interactions, such as integrating public transit schedules with road traffic data, enabling planners to optimize travel demand distribution~\cite{chalkidis_llama_2024}.  

In urban contexts, generative AI can predict congestion hotspots by analyzing heterogeneous data sources~\cite{de2023llm}, including weather conditions, socioeconomic indicators, and infrastructure utilization. For regional transportation planning, models extend predictions to assess long-term impacts of new policies, such as road tolls~\cite{chander2024advancing} or transit subsidies~\cite{yu2024large}, on travel demand and modal preferences. By providing accurate forecasts, AI enables planners to proactively address challenges like peak congestion, carbon emissions, and infrastructure bottlenecks.  

\textbf{Challenges.} Despite its potential, generative AI in predictive tasks encounters significant hurdles. 
\begin{itemize}[leftmargin=*]
	\item \textit{Local Nuances and Data Biases:}  A primary challenge lies in capturing \textit{local nuances} in transportation systems, such as region-specific travel behavior, socioeconomic variability, or infrastructure design, which can result in regionally biased or inaccurate forecasts~\cite{wang2024large}. Moreover, reliance on historical data introduces the risk of perpetuating existing biases, particularly those favoring well-documented regions or populations, while neglecting underserved or rural areas~\cite{bang2024measuring}.
	\item \textit{Real-time Adaptability:} Generative AI also struggles with \textit{real-time adaptability}, where rapidly changing conditions, such as weather disruptions, special events, or unexpected infrastructure failures, require models to dynamically update predictions. The computational cost of processing real-time data and the integration of multi-modal inputs, such as sensor data, crowd-sourced information, and policy changes, further complicate implementation~\cite{liu2024poliprompt}. 
	\item \textit{Explainability and Trust:} The lack of explainability in generative AI models, especially in deep learning models~\cite{zhang2024regexplainer}, is a significant hurdle for their adoption in high-stakes decision-making processes within transportation planning. This opacity makes it difficult for planners to understand and trust the reasoning behind AI-generated forecasts.
\end{itemize} 

To address these challenges, future research must focus on developing adaptive, real-time models that integrate diverse, high-quality datasets while accounting for dynamic system changes. Incorporating domain-specific knowledge and uncertainty quantification frameworks~\cite{liu2025mcqa, da2025understanding, da2024llm} can enhance robustness and interpretability, enabling planners to make informed decisions with confidence in AI-driven forecasts~\cite{chen2025uncertainty}. Collaborative efforts between AI researchers, transportation experts, and policymakers will be essential to ensuring equitable and reliable predictive solutions.

\subsection{Generative Tasks for Data Synthesis and Scenario Generation}\label{generativeTask}

\textbf{Definition.} Generative tasks focus on creating synthetic datasets, simulating hypothetical scenarios, or expanding incomplete data to enable robust transportation analysis~\cite{yu2024synthetic}. These tasks are particularly valuable when real-world data is limited due to collection costs, privacy concerns, or rare-event occurrences.

Generative AI models excel at \textit{synthetic data generation}, filling critical gaps in real-world datasets. For example, GANs or diffusion models can simulate realistic OD matrices for congestion analysis under disruptive events like natural disasters or large-scale public events~\cite{liu2024generative, chen2023review, yang2021bim}. Such synthetic datasets allow planners to test infrastructure resilience, policy outcomes, or urban mobility shifts without relying solely on empirical observations.

In addition to data synthesis, generative AI supports \textit{scenario generation}, enabling planners to evaluate multiple infrastructure or policy alternatives. For example, models can simulate transit-oriented developments~\cite{ma2018sustainable}, multimodal route designs~\cite{wang2022large}, or carbon impact assessments~\cite{heinonen2020spatial} under varying conditions. 
These capabilities are particularly useful for evaluating trade-offs between competing goals, such as minimizing travel time versus reducing emissions~\cite{ moore2010role}. 
Recent advancements demonstrate how human-guided generative models can enhance scenario diversity while aligning with stakeholder priorities in urban mobility planning~\cite{wang2023human,hu2024dual}. 
Furthermore, research leveraging conditional generation has shown promise in balancing various objectives (e.g., minimizing travel time, improving green coverage)~\cite{wang2021deep,wang2023automated}.
These approaches provide a robust and interactive foundation for comprehensive transportation analysis.
They enable decision-makers to simulate and compare policy outcomes systematically, supporting long-term sustainable transportation solutions.

\textbf{Challenges.} While generative AI offers transformative capabilities, ensuring that synthetic data and scenarios align with real-world behaviors is a significant challenge. Generated datasets often inherit biases from training data or produce unrealistic outputs, undermining their applicability to decision-making~\cite{yu2024large}. Robust validation frameworks that integrate domain expertise with statistical checks are essential to maintain reliability~\cite{gonzalez2024mitigating}. Adaptive techniques, such as iterative fine-tuning with real-world data or human-in-the-loop systems, can further refine outputs to align with practical constraints like budget, infrastructure capacity, and environmental regulations. Additionally, generative AI must address generalizability across diverse transportation contexts, requiring representative training datasets and uncertainty quantification tools to build confidence in their application. Addressing these challenges through technical innovation and interdisciplinary collaboration is critical to unlocking the potential of generative AI for sustainable, equitable transportation systems.

\subsection{Simulation Tasks for Traffic Dynamics }\label{simulationTask}
\textbf{Definition.} Simulation tasks involve modeling complex transportation systems, behaviors, and interactions under various conditions~\cite{de2021simulation,chen2018auto}. Generative AI enhances traditional simulation methods by enabling adaptive, high-fidelity representations of traffic flows~\cite{jiang2023generative}, vehicle coordination, and multimodal interactions.

In large-scale traffic networks, generative models simulate the effects of interventions such as road closures~\cite{lee2024driving}, traffic signal timing adjustments, or congestion pricing. By integrating real-time sensor data and historical patterns, AI-driven simulations provide actionable insights for traffic management and infrastructure planning~\cite{samaei2023comprehensive}.

\textbf{Challenges.} Despite their transformative potential, generative AI-driven simulations face several challenges that limit their broader applicability. 
\begin{itemize}[leftmargin=*]
	\item \textit{Dynamic behaviors:} A significant hurdle lies in accurately capturing the \textit{dynamic interplay} between vehicles, pedestrians, and infrastructure within transportation systems. Real-world traffic behavior is influenced by diverse factors, such as driver psychology, weather conditions, and socio-economic variability, which generative models often oversimplify or fail to incorporate effectively~\cite{jiang2023generative}. This can lead to unrealistic simulations that do not fully represent the complexities of mixed-autonomy systems or multimodal interactions.
	\item \textit{Computational cost:} Another critical challenge is the \textit{computational cost} associated with running high-resolution, real-time simulations at scale. Generative AI models integrated into \textit{digital twins} or \textit{predictive traffic systems} require significant computational resources to process diverse data streams, such as real-time sensor feeds, GPS traces, and dynamic OD matrices~\cite{samaei2023comprehensive}. While digital twins offer an opportunity to mirror real-world transportation systems for testing interventions, their reliance on generative AI magnifies the resource demands, particularly for urban-scale or multimodal networks. Developing lightweight and computationally efficient models is crucial to for scalability constraints.
	\item \textit{Generalization across scenarios} remains another concern. Models trained on one city's traffic patterns may struggle to adapt to different geographic, demographic, or cultural contexts, reducing their transferability and utility for global transportation challenges~\cite{lee2024driving}. Additionally, reliance on historical data introduces the risk of perpetuating biases, potentially excluding underrepresented regions or populations from benefiting fully from generative AI-driven simulations.
\end{itemize}

To address these challenges, future research must prioritize developing resource-efficient algorithms tailored for integration with digital twins and predictive systems. Enhancing scalability through distributed computing and adaptive model architectures can help address computational constraints. In addition, incorporating regional calibration techniques and domain knowledge will ensure broader applicability across diverse contexts.

\subsection{Trustworthiness in Generative AI-Based Transportation} \label{trustworthy}
Diffusion models and large language models have emerged as state-of-the-art generative frameworks, transforming numerous aspects of human life, including the transportation sector. Their applications in transportation range from autonomous vehicle navigation, traffic flow optimization, and intelligent infrastructure management to real-time congestion forecasting and transit planning. However, their widespread adoption has also unveiled inherent risks, raising significant concerns regarding their trustworthiness~\cite{fan2023trustworthiness, Li_Ji_Wu_Li_Qin_Wei_Zimmermann_2024, li2024dpudynamicprototypeupdating}. Ensuring generative AI's safe and ethical deployment in transportation requires addressing critical issues across six key dimensions: privacy, security, fairness, responsibility, explainability, and reliability.

\noindent \textbf{Privacy.}  
Ensuring privacy-preserving AI models has become a global priority \cite{carlini2023}, as privacy breaches can erode user trust, lead to malicious exploitation, and violate regulatory policies. In transportation, generative AI models are increasingly used in personalized travel planning, ride-sharing services, and vehicle-to-infrastructure communications, all of which involve sensitive user data such as location histories, travel patterns, and behavioral preferences. Diffusion Models and Large Language Models are particularly susceptible to privacy leakage \cite{Carlini2021} since they inherently capture the statistical properties of their training data. Unauthorized access to this information can expose sensitive details, such as frequent travel routes or home addresses, making it imperative to design models that safeguard user data within transportation systems.

\noindent \textbf{Security.}  
The robustness of Diffusion Models and LLM against malicious manipulation is crucial for their safe deployment, especially in transportation, where security breaches can lead to life-threatening consequences. Two prevalent attack vectors are adversarial attacks \cite{pmlr-v202-liang23g} and backdoor attacks \cite{Chen2023}. Adversarial attacks exploit vulnerabilities by introducing imperceptible perturbations to input data, deceiving the model into making incorrect predictions. In autonomous driving systems, such perturbations could cause a vehicle to misinterpret traffic signs or road conditions, leading to accidents. Backdoor attacks, on the other hand, embed concealed triggers within the model, allowing adversaries to control its behavior upon activation. For example, malicious actors could exploit these vulnerabilities to manipulate traffic signal control systems, resulting in traffic congestion or collisions. These security threats can compromise model integrity, leading to unpredictable and potentially harmful consequences in transportation networks.

\noindent \textbf{Fairness.}  
As generative models increasingly influence real-world decision-making, including in transportation applications like dynamic pricing in ride-sharing, traffic management, and route optimization, it is essential to uphold fairness and prevent algorithmic biases that may disproportionately affect certain demographic groups. For instance, biases in traffic prediction models may result in suboptimal route recommendations for underserved communities or unfair prioritization in traffic management systems. These models should align with ethical standards to avoid reinforcing societal prejudices and discrimination. However, biases embedded in training data often manifest in AI-generated content \cite{Wallace2019Triggers, Li2023MM}, resulting in inequitable outcomes and furthering social disparities. Addressing fairness in transportation-focused generative AI applications is crucial to ensuring equitable access and reliable services for all user groups.

\noindent \textbf{Responsibility.}  
Responsibility in AI-driven transportation systems encompasses three key aspects: identifiability, traceability, and verifiability. Ensuring that AI-generated transportation policies, infrastructure recommendations, and predictive analyses are clearly distinguishable from human-generated content is crucial to preventing misinformation. Embedding traceability markers, such as watermarks or metadata, allows planners and policymakers to verify the origin and credibility of AI-generated insights. Additionally, verifiability in AI-driven decision-making enhances transparency, ensuring that traffic optimization strategies, demand forecasts, and infrastructure proposals align with real-world transportation objectives and expert assessments.

\noindent \textbf{Explainability.}
Explainability is essential in AI-driven transportation planning, allowing stakeholders to understand and validate the reasoning behind AI-generated outputs. Without clear explanations, planners may struggle to interpret why a model recommends certain congestion mitigation measures, transit expansions, or traffic signal adjustments. Feature attribution techniques, such as SHAP~\cite{antwarg2021explaining}, help identify which factors most influenced AI predictions, improving transparency in AI-driven decision-making. Chain-of-Thought reasoning~\cite{wei2022chain} further enhances interpretability by structuring AI responses into sequential, human-readable justifications, making AI outputs more accessible to policymakers and transportation agencies.

\noindent \textbf{Reliability.}
Reliability in AI-based transportation models is critical for ensuring that generative AI systems produce consistent, trustworthy predictions. AI models often operate in conditions of incomplete or uncertain data, making uncertainty quantification a crucial component of reliability~\cite{young2024flexible}. In Large Language Models (LLMs), uncertainty quantification has gained attention due to the inherent variability in generated text and reasoning outputs\cite{lin2023generating, kuhn2023semantic,da2024llm}. Self-consistency decoding, for instance, prompts LLMs multiple times with slightly perturbed inputs to evaluate the consistency of generated outputs\cite{wang2022self}. If multiple responses converge toward the same answer, the model exhibits higher confidence; if responses vary widely, the uncertainty level is high. 
Retrieval-Augmented Generation (RAG)~\cite{lewis2020retrieval} with confidence scoring is another method that improves uncertainty quantification in LLMs~\cite{hasegawa2024rag}. Instead of relying solely on internal model knowledge, RAG-enhanced models retrieve external transportation data sources, such as real-time congestion reports or infrastructure databases, to refine AI-generated recommendations. By assigning confidence scores to retrieved information, planners can differentiate between high-certainty and low-certainty predictions in transportation system optimization. Without robust uncertainty-aware methodologies, AI-driven transportation solutions risk overconfidence in unreliable predictions, leading to misallocated resources and ineffective mobility policies.

\section{Technical Foundations for Generative AI Applications in Transportation Planning} \label{Computer-Science}
%Technical Foundations for LLM Adaptation in Political Science
%Original Title: Computational Approaches for Advancing LLMs in Political Science

\subsection{Dataset Preparation}
\label{Benchmark}

\begin{table*}[htbp]
	\footnotesize
	\centering
	\caption{Benchmark datasets for generative AI applications in transportation planning.}
	\label{tab:transportation-datasets}
	\resizebox{0.8\linewidth}{!}{
		\begin{tabularx}{\textwidth}{p{3cm} p{4.5cm} X}
			\toprule
			\centering \textbf{Data Type} & \textbf{Description} & \textbf{Use in Transportation Planning} \\
			\midrule
			\centering Census Transportation Planning Products (CTPP)~\cite{seo2017ctpp,ctpp} & Data from the American Commuter Survey (ACS) designed to understand commuter information such as where people commute to and from, and how they get there. & Used to inform travel demand models and forecast future travel trends and infrastructure investments. Includes data on commute origins, destinations, mode choice, and work locations. \\
			\centering Census Bureau Data~\cite{censusdata,simon2022census,freiman2017data} & Data from the U.S. Census Bureau, including population, employment, income, and housing information. & Provides essential demographic and economic data to inform long-term transportation planning and investment decisions. Includes population data, employment statistics, and housing data. \\
			\centering National Household Travel Survey (NHTS)~\cite{hu2004summary,santos2011summary,mcguckin2018summary} & Data on travel and transportation patterns in the United States, including trip purposes, modes of transport, and travel time. & Provides insights into travel behavior, informing the development of transportation infrastructure and services. Includes trip purpose data, vehicle type data, and travel time data. \\
			\centering Traffic Count Data~\cite{da2024open} & Data on the volume of vehicle travel on highways and roads, including vehicle counts, travel speeds, and vehicle classifications. & Used to calibrate and validate travel demand models, show growth trends, and guide infrastructure development decisions. Includes vehicle count data, travel speed data, and vehicle class data. \\ 
			\centering Transit Passenger Surveys~\cite{ji2015transit,zhang2018different} & Data collected from transit riders regarding their demographics, travel patterns, and preferences. & Used to improve public transportation systems, optimize routes, and address service gaps. Includes transit ridership surveys, travel time, and transfer data. \\
			\centering Probe Vehicle Data from Vendors (e.g., INRIX, Streetlight)~\cite{molloy2023mobis,yang2020guidelines,mavromatis2022dataset} & Data from GPS-enabled devices and mobile apps capturing vehicle movements. & Used for traffic flow analysis, congestion monitoring, and route optimization. Includes vehicle speed, travel time, and route choices. \\
			\centering High-Resolution Vehicle Trajectory~\cite{zachar2023visualization,moers2022exid} & Detailed vehicle paths from connected vehicles and GPS devices, capturing second-by-second movements. & Helps analyze microscopic traffic behavior, signal timing optimization~\cite{chen2024syntrac}, and bottleneck identification. Includes second-by-second GPS traces, turn movements, and acceleration/deceleration rates. \\
			\centering Connected and Automated Vehicle (CAV) Data~\cite{kang2019test,sun2020scalability} & Data from CAV systems, including telemetry and communication logs. & Enables predictive traffic management, safety analysis, and infrastructure planning. Includes lane usage, vehicle speed, travel time, and route choices. \\
			\centering Crowd-Sourced Platform Data (e.g., Waze, Google Maps)~\cite{wazeforcities} & User-generated data from apps and social platforms. & Tracks real-time traffic incidents, congestion hotspots, and public sentiment. Includes traffic incident reports, delay alerts, and congestion levels. \\
			\centering Geospatial Data~\cite{cheng2017remote,stewart2022torchgeo} & Geographic data representing road networks, infrastructure details, and land-use patterns. & Essential for creating maps, analyzing network performance, and planning infrastructure developments. Includes road networks, infrastructure locations, and zoning data. \\
			\centering Demographic Data~\cite{zhang2024netmob2024} & Population density, socioeconomic status, geographic distribution, and travel behavior data. & Helps understand community transportation needs, predict demand, and optimize routes and services for diverse groups. Includes population distribution, income levels, and vehicle ownership rates. \\
			\centering Environmental Data~\cite{gibert2018environmental} & Data related to environmental conditions, such as weather, emissions, and noise levels. & Used for evaluating environmental impacts of transportation projects and planning sustainable, eco-friendly solutions. Includes weather patterns, carbon emissions, and air quality indices. \\
			\centering Freight Movement Data~\cite{shoman2023review} & Data from logistics providers, ports, and supply chains, including freight volume, transit times, and routes. & Supports planning for freight infrastructure, logistics, and optimizing transportation networks for goods movement. Includes freight flow data, port data, and logistics timing. \\
			\centering Social Media Data~\cite{yao2021twitter} & Public posts, comments, and feedback from platforms like Twitter (X), Facebook, and Instagram. & Can be used to analyze public concerns, traffic incidents, or service disruptions, as well as to track public perceptions. Includes tweets, Facebook posts, and Instagram comments. \\
			\centering Accident and Safety Data~\cite{huang2023tap} & Data on traffic accidents, fatalities, and incidents from police reports, insurance data, and sensors. & Used to predict accident hotspots, improve safety measures, and optimize emergency response strategies. Includes accident reports, insurance claims, and sensor data on accidents. \\
			\centering Transit Data~\cite{widmer2023ztbus} & Data from public transportation systems, including ridership, schedules, delays, and vehicle conditions. & Helps improve public transportation efficiency, optimize routes, and ensure coverage in underserved areas. Includes bus ridership data, subway schedules, and vehicle performance data. \\
			\bottomrule
		\end{tabularx}
	}
\end{table*}

\subsubsection{Benchmark Datasets}
To meet the specific demands of transportation planning and research, a variety of benchmark datasets have been developed to evaluate generative AI models on tasks such as traffic forecasting, demand modeling, infrastructure planning, policy evaluation~\cite{da2024probabilistic}, and public sentiment analysis. These datasets are designed with domain-specific criteria to ensure that AI outputs are relevant and applicable to real-world transportation scenarios. A comprehensive list of these datasets, along with their respective tasks and characteristics, is summarized in Table~\ref{tab:transportation-datasets} to facilitate reference and comparison.

\noindent \textbf{Traffic Forecasting and Demand Modeling Datasets.} Traffic forecasting and demand modeling are foundational for understanding transportation dynamics and optimizing mobility systems. The METR-LA dataset~\cite{wang2023correlated} contains traffic data from Los Angeles, collected from 207 sensors over several months, providing a resource for evaluating AI models on spatiotemporal forecasting tasks. Similarly, the PEMS-BAY dataset~\cite{chen2020multi, wu2019graph} includes traffic flow information from the San Francisco Bay Area, enabling the analysis of regional mobility patterns.

For demand modeling, OpenStreetMap and GTFS (General Transit Feed Specification) datasets~\cite{wong2013leveraging} provide multimodal transit network details and schedules, supporting generative AI applications in routing, schedule optimization, and network design. The NYC Taxi Dataset~\cite{nyctaxi2020} tracks millions of taxi trips across New York City, providing high-resolution data for demand prediction and dynamic pricing analysis. These datasets serve as benchmarks for training and evaluating generative models in traffic flow and travel demand scenarios.

\noindent \textbf{Public Sentiment and Engagement Datasets.}
Understanding public sentiment is crucial for transportation policy-making and project evaluation. The Public Opinion and Sentiment Dataset (POSD)~\cite{wong2013leveraging} aggregates feedback from community surveys, online forums, and social media about transportation projects. This dataset helps generative AI models analyze public preferences and predict responses to proposed infrastructure changes.
The Global Public Transit Dataset (GPTD)~\cite{ngsim2024opendata} focuses on user reviews and social media posts about public transit systems worldwide. It allows for sentiment analysis at a granular level, providing insights into specific concerns such as punctuality, cleanliness, and affordability. Generative AI trained on these datasets can provide actionable recommendations to improve public transit experiences.

\noindent \textbf{Infrastructure Planning and Resilience Datasets.}
Generative AI can use datasets tailored to infrastructure design and resilience planning. The OpenStreetMap (OSM) Road Networks~\cite{memduhouglu2024enriching} dataset includes detailed maps of global road networks, enabling AI to generate infrastructure designs that balance connectivity and accessibility. The Resilient Infrastructure Dataset (RID)~\cite{jeffers2018analysis} contains data on infrastructure performance under extreme weather events, supporting the generation of adaptive infrastructure strategies.
Additionally, the UrbanSim dataset~\cite{fan2024does} provides city-scale data on land use, population density, and transportation networks, allowing generative AI to simulate the impacts of zoning and infrastructure policies on urban mobility.

\noindent \textbf{Scenario Simulation and Optimization Datasets.}
Generative AI is frequently used for scenario simulation and optimization in transportation systems. The SUMO (Simulation of Urban MObility) dataset~\cite{behrisch2011sumo} provides multimodal transportation simulations, enabling AI to generate efficient and sustainable urban mobility plans. The Traffic Simulation Dataset (TSD)~\cite{zhang2022traffic} includes synthetic traffic data for exploring congestion management strategies under varying conditions.
The Infrastructure and Emissions Dataset (IED)~\cite{wang2023infrastructure} integrates traffic flow data with carbon emission metrics, allowing generative AI models to optimize transportation networks while minimizing environmental impacts.

These benchmark datasets provide a robust foundation for generative AI applications in transportation planning. From forecasting travel demand and analyzing public sentiment to optimizing infrastructure resilience and detecting misinformation, these datasets support a wide range of use cases. By leveraging domain-specific datasets, generative AI models can deliver actionable insights and equitable solutions tailored to the evolving needs of modern transportation systems.

\subsubsection{Dataset Preparation Strategies}
\label{dataset-preparation}

Dataset preparation is a crucial step in adapting generative AI for downstream transportation planning applications~\cite{yu2024makes}. As the application of generative AI in transportation is still emerging, publicly available benchmark datasets are often limited in scale and scope. Developing effective transportation-specific datasets requires careful consideration of domain-specific strategies, while drawing insights from adjacent fields such as time-series forecasting, infrastructure modeling, sentiment analysis, and simulation tasks~\cite{lin2024designing,wagner2024power}. These strategies ensure alignment with key transportation tasks, including traffic prediction, demand modeling, policy evaluation, and infrastructure optimization.

Transportation planning is inherently local in nature due to jurisdictional boundaries, varied demographic profiles, and unique traffic conditions specific to each region. For example, driving behaviors and transportation priorities in Los Angeles differ significantly from those in New York City, rural Midwest regions, or urban centers in Asia. Consequently, building datasets that are scalable across jurisdictions requires addressing substantial variability in traffic conditions, demographic distributions, and policy environments. Reconciling local variations, merging datasets from different regions, and ensuring inter-jurisdictional consistency form a significant portion of the dataset preparation effort.

To ensure the quality and reliability of generative AI models in transportation planning, datasets must meet the following key requirements:
\begin{itemize}[leftmargin=*]
	\item \textit{Completeness:} Data must encompass diverse transportation conditions, such as peak versus off-peak hours, weekdays versus weekends, varying weather conditions, and cross geographic and jurisdictional boundaries. For example, datasets for traffic flow analysis should include representative samples from all these scenarios to avoid biased predictions or suboptimal planning outcomes. Public agencies often operate independently within their defined regions, which means reconciling diverse datasets requires substantial coordination, synthesis, and agreement. Diverse, high-quality data can facilitate collaboration among agencies, researchers, and private stakeholders to enable actionable insights and reliable planning outcomes. Addressing geographical and behavioral differences, such as contrasting driving habits in urban and rural areas, is essential for comprehensive data coverage.
	\item \textit{Accuracy:} High-quality data ensures that AI models can produce reliable outputs. Real-time traffic data, for instance, must be consistently updated, while geospatial data must accurately reflect current road networks, land use, and infrastructure layouts. Errors in the data can propagate through AI models, leading to flawed conclusions.
	\item \textit{Consistency:} Harmonizing datasets from different sources is critical. For instance, demographic data from varying agencies or surveys must align with traffic flow observations to avoid conflicting outputs. Ensuring consistency across datasets facilitates seamless model training and deployment, especially when combining data from multiple jurisdictions.
	\item \textit{Granularity:} Transportation applications often require data at specific levels of detail. For example, traffic flow analysis may need second-by-second GPS traces, while long-term policy evaluations might focus on broader demographic trends. Ensuring datasets have appropriate granularity enables tailored applications while avoiding computational inefficiencies.
\end{itemize}

\subsubsection{Broad Source of Dataset Collection}\label{broadsource}
A primary strategy for dataset preparation involves collecting data from publicly available transportation sources, such as traffic sensors, GPS systems, government reports, and user-generated feedback. For example, traffic flow data from sensor networks or ride-hailing platforms can be adapted for congestion prediction tasks, while infrastructure reports can serve as input for generative models that simulate adaptive urban designs.
In sentiment analysis applications, datasets such as public transit reviews, social media posts, and community surveys provide critical insights into user satisfaction and preferences. For instance, the Global Public Transit Dataset~\cite{zhang2018different} and similar resources aggregate user feedback from social platforms to analyze commuter perceptions and identify service bottlenecks. Such datasets can be annotated to support tasks like sentiment detection, network optimization, and feedback-driven policy recommendations.
For policy evaluation and impact analysis, historical records of transportation policies, government documents, and infrastructure development data serve as foundational inputs. Datasets like the CBLab Dataset~\cite{liang2023cblab} combine policy details with socioeconomic variables, enabling generative AI to model the short- and long-term impacts of transportation interventions, such as congestion pricing or multimodal system integration.

\vspace{1mm}
\noindent \textbf{Data Annotation and Labeling.}
Once collected, transportation datasets often require annotation and labeling to align with specific tasks. For instance, in traffic analysis, congestion levels, travel times, and accident severity may be labeled to train generative AI models for accurate prediction and scenario generation~\cite{zhang2024generative}. Similarly, infrastructure datasets may be annotated with resilience scores, environmental impacts, or maintenance needs to support generative tasks, such as simulating infrastructure upgrades~\cite{andreoni2024enhancing}.
Annotation strategies also play a key role in sentiment and engagement analysis. For example, public feedback from transit networks can be labeled to categorize positive or negative sentiments toward specific transportation services, such as reliability, accessibility, or cleanliness~\cite{das2023leveraging}. High-quality annotations ensure that generative AI models capture nuanced user preferences and generate actionable recommendations for transportation planners.

Annotation can be conducted through different approaches. These methods range from fully manual labeling~\cite{tan2024large}, where annotation experts review and label the data by hand, to semi-automated processes that use algorithms to assist with labeling~\cite{huang2024selective}, with experts intervening as needed. In fully automated labeling, LLMs or other automated systems can handle the labeling work entirely, followed by a quality check~\cite{ming2024autolabel}. Each method has its trade-offs among accuracy, scalability, and manual effort required.

\vspace{1mm}
\noindent \textbf{Ensuring Data Diversity and Representativeness.}
To ensure generalizability, transportation datasets must represent diverse geographies, demographic groups, and environmental conditions. For instance, combining urban, suburban, and rural datasets enables generative AI models to address transportation challenges across different spatial contexts. In addition, datasets incorporating varied socioeconomic conditions, such as underserved communities or emerging economies, help mitigate biases and promote equitable solutions.

\vspace{1mm}
\noindent \textbf{Data Augmentation for Generative Tasks.}
Data augmentation techniques enhance dataset size and diversity, especially when real-world data is scarce or incomplete. For traffic forecasting, techniques such as synthetic trajectory generation and perturbation modeling can create realistic yet varied travel patterns. In infrastructure optimization, generative models can simulate alternative urban layouts, enabling planners to evaluate multiple design scenarios under different constraints, such as cost, population density, and environmental impact.

To further illustrate how these strategies apply to practical scenarios, we now introduce three examples of dataset preparation tailored for generative AI applications in transportation planning. Each example demonstrates how researchers effectively leverage generative AI to address key challenges in transportation data curation and annotation:

\vspace{1mm}
\noindent \textbf{1. Developing a Dataset for Bias Mitigation in Transportation Models}
For tasks requiring equitable transportation planning, constructing a balanced dataset involves curating diverse data sources that represent different geographical areas, socioeconomic conditions, and demographic groups. For example, to address bias in generative AI outputs, datasets can include multimodal transit usage patterns, ride-hailing data, and urban-rural travel surveys. These datasets should be annotated with indicators such as accessibility, affordability, and network efficiency. Annotation can combine manual input from transportation experts and automated detection tools to identify inequities in system performance or user experience. The goal is to create datasets that enable generative AI models to recognize and address biases, ensuring outputs are inclusive and equitable across diverse communities.

\vspace{1mm}
\noindent \textbf{2. Automated Annotation Using Generative AI: Example in Infrastructure Analysis.}
Infrastructure analysis is a critical task in transportation planning, requiring extensive data on road networks, transit systems, and environmental conditions. Generative AI can be employed to automate the annotation of datasets like OpenStreetMap~\cite{bennett2010openstreetmap}, which provides detailed global road network data. For instance, generative models can annotate road segments with features such as congestion levels, maintenance status, and connectivity to multimodal systems. Fine-tuning these models on smaller, manually annotated datasets enables accurate classification of infrastructure attributes and potential bottlenecks. This automated annotation process accelerates the analysis of large-scale infrastructure data, providing planners with actionable insights to optimize network designs and prioritize maintenance.

\vspace{1mm}
\noindent \textbf{3. Generating Synthetic Transportation Datasets Using Generative AI.} 
The limited availability of diverse transportation datasets, due to privacy concerns or data collection challenges, makes synthetic dataset generation a valuable solution. For example, in traffic forecasting tasks, generative AI models can simulate hypothetical traffic scenarios based on historical sensor data and projected urban growth patterns. By training generative AI on existing traffic datasets, researchers can produce synthetic datasets that mimic real-world conditions, including variations in congestion, weather, and commuter behaviors. These synthetic datasets can enhance model robustness by exposing generative AI to diverse scenarios, enabling more accurate predictions and simulations for urban mobility planning.

\subsection{Fine-Tuning Generative AI for Transportation Planning} \label{fine-tuning}

This section presents the general technique of fine-tuning generative AI models, then demonstrates how the technique is adopted in the specific context of OD Demand Calibration in transportation planning.

\subsubsection{General Steps in Fine-Tuning Generative AI Models}
Fine-tuning refers to adapting a pre-trained generative model to better suit a specific downstream task. The process generally involves:
\begin{enumerate}[label=\arabic*., leftmargin=*]
	\item \textbf{Data Preprocessing:} Aligning input-output data pairs through normalization, formatting, and noise filtering to create structured training examples.
	\item \textbf{Model Adaptation:} Using parameter-efficient methods such as LoRA or prefix-tuning to modify a small subset of weights.
	\item \textbf{Training:} Performing loss minimization on structured data using strategies like gradient accumulation and mixed-precision training, while validating on held-out subsets.
	\item \textbf{Prompt Engineering:} Designing task-specific prompts to condition the model outputs in alignment with domain requirements.
	\item \textbf{Evaluation:} Measuring performance using relevant metrics such as RE (Relative Error) and RMSE.
\end{enumerate}

\subsubsection{Specific Problem Setting: OD Demand Calibration}
OD (Origin-Destination) Demand Calibration estimates travel demand between regions by adjusting OD matrices to match observed traffic counts. Accurate OD matrices are essential for effective infrastructure planning and traffic management. 

\paragraph{Domain-Specific Dataset} The OpenTI dataset~\cite{da2024open} supports OD calibration with rich traffic flows, network structures, and travel demands. It provides:
\begin{itemize}
	\item \textbf{Input:} Raw OD matrices and observed link-level traffic counts.
	\item \textbf{Output:} Calibrated OD matrices aligned with observed traffic patterns across time intervals.
\end{itemize}

\paragraph{Adopting Fine-Tuning Steps for OD Calibration}
Each general step is instantiated in this application as follows:

\textbf{1. Data Preprocessing:}
Preprocessing OpenTI data ensures consistency between inputs (raw OD matrices and observed traffic counts) and outputs (calibrated OD matrices). The steps include:
(1) Normalization: Scaling OD demand values to ensure uniform ranges across datasets.
(2) Matrix Formatting: Aligning OD matrices with observed traffic counts to maintain input-output coherence.
(3) Noise Reduction: Filtering out inconsistencies or redundant data points to streamline the training process.
The preprocessed data pairs provide well-structured examples for training generative AI models to produce calibrated OD matrices.

\textbf{2. Model Adaptation:} Apply LoRA~\cite{wu2024dlora} or prefix-tuning~\cite{meloux2024novel} to an LLM such as Llama3.1-8B.

\textbf{3. Training Process:} Utilize GPU clusters, gradient accumulation~\cite{nabli2024acco}, and mixed-precision training~\cite{guan2024aptq}. Minimize loss between generated and ground-truth calibrated OD matrices.

\textbf{4. Prompt Engineering:} Craft prompts that direct the model to adjust demand values while preserving traffic consistency:

\begin{tcolorbox}[colback=gray!4!white, colframe=gray!75!green, title=\textbf{Prompt Engineering Examples for OD Calibration}] \centering \faLightbulbO: Prompt \hfill \raggedright \rule{13cm}{1pt} \linespread{1.25} \selectfont
	\small
	\faLightbulbO \ \ding{182}: Given the raw OD matrix and observed traffic counts for the following network, generate a calibrated OD matrix that minimizes the error between predicted and observed flows.
	
	\faLightbulbO \ \ding{183}: Adjust the demand values in the provided OD matrix to align with the traffic counts observed during peak hours, ensuring network flow consistency.
	
	\faLightbulbO \ \ding{184}: Based on the input OD matrix and flow observations, calibrate the OD demand such that the relative error across all links is minimized.
	
	\faLightbulbO \ \ding{185}: Using the raw OD matrix and observed data, generate a revised demand matrix that reflects travel demand during weekend traffic conditions.
	
\end{tcolorbox}

\textbf{5. Evaluation:}
\begin{itemize}
	\item \textit{Relative Error (RE):} Compares predicted vs observed counts.
	\item \textit{RMSE:} Measures deviation from ground-truth OD matrices.
\end{itemize}

\paragraph{Expected Outputs and Impact}
After fine-tuning on the OpenTI dataset, the generative AI model is expected to perform effectively across several OD calibration tasks:

\begin{itemize}[leftmargin=*]
	\item \textit{Accurate Demand Calibration:} The model should generate OD matrices that closely align with observed traffic counts, reducing errors and improving demand estimation accuracy. For example, the model might adjust an input matrix to account for increased north-south traffic flow during morning peak hours. To prove the effectiveness of fine-tuning LLMs, we fine-tune a Llama3.1-8B model on OpenTI dataset in \cref{caseStudy} and the results show that a 8B model could outperform a 70B LLM, indicating the great performance gain from fine-tuning.
	\item \textit{Adaptation to Diverse Traffic Scenarios:} Due to the commonsense knowledge from base LLM model, the fine-tuned model generalizes to unseen networks or varying conditions, such as adjusting demand for weekend traffic, construction detours, or adverse weather impacts quickly. 
	\item \textit{Structured and Interpretable Outputs:} The model outputs are structured OD matrices with clear documentation of adjustments, ensuring interpretability for transportation planners. Based on the design of training dataset, the interpretability could even be enhanced by the clear explanation in the training data. Outputs could balance demand realism with network flow consistency.
\end{itemize}

Fine-tuning generative AI models using the OpenTI dataset enables precise, efficient OD demand calibration, addressing a critical challenge in transportation planning. Through systematic preprocessing, parameter-efficient fine-tuning, and tailored prompt engineering, the model can produce calibrated OD matrices that improve the demand estimation accuracy, reduce errors, and enhance real-world applicability. We show detailed evaluation in \cref{caseStudy} and the result demonstrates the potential of generative AI to optimize traffic management systems and support data-driven decision-making in modern transportation networks.

\subsection{Zero-Shot Inference with Generative AI for Transportation Accessibility} \label{zero-shot}

This section presents the general framework of Zero-Shot Learning (ZSL) using generative AI, and demonstrates how it is applied to estimating city-level public transportation accessibility, particularly through People Near Transit (PNT) scores.

\subsubsection{General Methodology: Zero-Shot Learning (ZSL) with LLMs}
ZSL enables a pre-trained LLM to perform domain-specific tasks without any fine-tuning or labeled examples. This is achieved solely through carefully constructed prompts. For urban and transportation studies, ZSL offers a scalable and cost-effective tool for analyzing cities lacking structured datasets.
Key Features of ZSL are: {(1)~No training required:} Tasks are performed via prompt design without additional data or tuning. {(2)~Wide applicability:} General knowledge encoded during pretraining is leveraged for domain-specific inference. {(3) Interpretability:} Prompts and outputs are transparent and human-readable, supporting validation.

\subsubsection{Specific Problem Setting: Estimating PNT in Global Cities}
People Near Transit (PNT) measures the proportion of a city’s population living within a defined distance (e.g., 500m, 1000m, 1500m) of transit stops. Directly obtaining such data requires detailed GIS and fieldwork, which are often unavailable.

\paragraph{Input Representation} To approximate PNT using LLMs, we convert spatial accessibility into qualitative prompts about urban conditions, including urban planning quality, road network conditions, public transportation extensiveness, and government support for transit development. 
Each factor is queried using a 0.0 to 10.0 scale, allowing the LLM to infer high-level characteristics of accessibility.

\paragraph{Adapting ZSL for Transportation Accessibility}
The most important step in applying ZSL is prompt engineering, i.e., Design prompts incorporating contextual information. An example prompt used in~\cite{li2024can} is shown below.

\paragraph{Outputs and Impact} Compare inferred responses with ground-truth PNT scores, where available, across spatial scales (500m, 1000m, 1500m). 
The responses generated by LLMs can be validated by directly correlating their outputs with ground-truth PNT (Proximity to Network Transit) scores \cite{verbavatz2020pnt}, where available. This approach provides a robust means of assessing the accuracy of predictions. Notably, this method demonstrates significant improvements compared to directly querying the LLMs for PNT scores without further calibration.
Table \ref{tab:pnt_result} summarizes the performance of LLM-generated predictions across varying spatial scales (500m, 1000m, and 1500m) in 85 global cities. The comparison covers three methods: Zero-Shot Learning (ZSL), direct queries, and using average PNT scores as predictions (AVG), and we find that ZSL consistently performs better than traditional methods. For cities without available transit data, the LLM's estimations provide a scalable and efficient approximation, which can support decision-making and hypothesis testing.

\begin{tcolorbox}[colback=gray!5!white, colframe=gray!75!green, title=\textbf{Estimating Public Transportation Accessibility in Global Cities}] \centering \faLightbulbO: Prompt \hfill \faRedditAlien: LLM Response \hfill \faQuestionCircleO: Explanation \raggedright \rule{13cm}{1pt} \linespread{1.25} \selectfont
	\small
	\faLightbulbO: Based on the information you have, estimate the level of public transportation accessibility for New York City. Consider factors like subway availability, bus services, and urban planning efficiency. Provide a score between 0 and 10.
	
	\faRedditAlien: The public transportation accessibility in New York City is likely very high, given its extensive subway system, comprehensive bus network, and urban design that prioritizes transit-oriented infrastructure. I estimate the accessibility score to be 9.5.
	
	\faQuestionCircleO: The response considers NYC's known public transit infrastructure, such as its subway network, which contributes to a high accessibility score. LLMs rely on general knowledge encoded during pretraining to generate such estimations. \end{tcolorbox}

\begin{table}[t!]
	\centering
	\begin{tabular}{@{}lcccccc@{}}
		\toprule
		\textbf{Category} & \textbf{Coverage}      & \textbf{ZSL} & \textbf{Direct Query}  & \textbf{AVG} \\ \midrule
		PNT (500m)                   & 85 global cities       & 10.54         & 14.27                  & 26.83            \\
		PNT (1000m)               & 85 global cities       & 13.22         & 19.41                   & 23.76            \\
		PNT (1500m)               & 85 global cities       & 13.98         & 19.92                  & 28.34            \\ \bottomrule
	\end{tabular}
	\caption{Summary statistics for PNT across different scales in 85 global cities.}
	\label{tab:pnt_result}
\end{table}

\paragraph{Comparison with Traditional Methods} Traditional approaches to estimating public transportation accessibility rely on geographic information systems (GIS) and extensive field surveys, which are resource-intensive~\cite{welch2019big}. ZSL, in contrast, extracts latent knowledge from LLMs using contextual prompts, offering a lightweight alternative for cities with limited data. As demonstrated in~\cite{jin2023large, da2024prompt, li2024can}, LLM-derived features often show strong correlations with real-world transportation metrics, highlighting the potential of ZSL as a viable estimation method.

\paragraph{Strengths and Challenges}
ZSL provides significant advantages for transportation planning:
It enables rapid scalability across global cities with minimal cost.
It can infer latent urban features (e.g., accessibility, traffic density) without training data. Additionally, ZSL introduces a novel user interface paradigm for interacting with transportation models, making advanced modeling tools more accessible and intuitive for practitioners. LLMs can simplify data queries, scenario simulations, and decision-making processes through conversational interfaces, lowering the barrier to entry for complex analysis and fostering broader adoption across the industry.
However, LLMs may exhibit uncertainties when tasked with cities for which they lack sufficient knowledge. Such cases often yield generic or inconsistent predictions. Therefore, validating LLM outputs against ground-truth data remains critical for ensuring reliability. 

\noindent \textbf{Conclusion.}
Zero-Shot Learning, as illustrated by PNT estimation, demonstrates the ability of LLMs to extract transportation accessibility insights across cities. By leveraging contextual prompt engineering, LLMs can provide valuable approximations in the absence of task-specific data, thereby offering scalable and efficient solutions for transportation studies. This approach opens new opportunities for applying generative AI to urban planning, particularly in under-researched or data-scarce regions~\cite{li2024whattell}.

\subsection{Inference with Generative AI: Few-Shot In-Context Learning}
\label{few-shot}

This section presents the general methodology of Few-Shot Learning (FSL) in generative AI and illustrates its application to traffic coordination in mixed-autonomy environments, using the CoMAL framework~\cite{yao2024comal} as an example.

\subsubsection{General Methodology: Few-Shot In-Context Learning (FSL)}
FSL enables large language models (LLMs) to generalize to new tasks by conditioning on a small number of examples included in a single prompt. It avoids the need for fine-tuning while still leveraging powerful generalization capabilities. 
Key features of FSL are (1)~Minimal supervision: Similar to zero-shot learning, no gradient updates or labeled training datasets are needed. (2)~{Flexible reasoning:} Enables task-specific behavior via few-shot examples and descriptive prompts. (3)~{Prompt-centered adaptation:} The model behavior is guided through structured examples within the input.

\subsubsection{Specific Problem Setting: Traffic Coordination in Mixed-Autonomy} In mixed-autonomy scenarios, connected autonomous vehicles (CAVs) and human-driven vehicles share the road. Coordinating these heterogeneous agents in real-time is challenging. The CoMAL framework uses LLMs and FSL to address this coordination challenge in scenarios such as highway merging, roundabouts, and intersections.

\paragraph{Adapting FSL for Traffic Coordination} The effectiveness of few-shot learning heavily relies on the selection of representative examples. For the task of Multi-Agent Traffic Optimization, CoMAL leverages LLM-based multi-agent systems to address the coordination of connected autonomous vehicles (CAVs) in mixed-autonomy traffic, where both autonomous and human-driven vehicles coexist. Few-shot examples are embedded in prompts to enable the LLM to infer strategies for optimizing traffic flow without extensive retraining. In the CoMAL framework, prompts include task descriptions, environment perception, and a few relevant examples of agent roles, driving instructions, and reasoning steps. 

\textbf{1. Few-Shot Prompt Design:} Carefully selected coordination examples are embedded into prompts, showing past scenarios of role assignments and motion strategies. In the prompt template titled `Few-Shot Prompt for Merging Traffic Coordination', few-shot examples provide scenarios with varying roles and motion plans. By generalizing from these examples, the LLM infers optimal coordination strategies for traffic merging, balancing acceleration, and deceleration to reduce congestion and shockwaves.

\begin{tcolorbox}[colback=gray!5!white, colframe=gray!75!green, title=\textbf{Few-Shot Prompt for Merging Traffic Coordination}] \centering
	\small
	\faLightbulbO: Prompt \hfill \faTags: Response \hfill \faMap: Context \hfill
	\raggedright \rule{13cm}{1pt} \linespread{1.25} \selectfont
	
	\faLightbulbO \ \ding{182}: "You are the brain of an autonomous vehicle (CAV). Your goal is to optimize traffic flow on a highway where vehicles are merging. Based on the following scenario, allocate roles (Leader, Follower) and generate motion plans for all CAVs. Avoid collisions and maintain safe headway distances."
	
	\textbf{Scenario Description:} Vehicles are merging onto the main road from an on-ramp. Your current position is $11.75 m$, speed $3.80 m/s$. $CAV_2$ is ahead at $15.64 m$, moving at $4.51 m/s$.
	
	\textbf{Example 1:}
	\faMap: "$CAV_1$ is ahead. $CAV_2$ should slow down to allow merging."
	\faTags: Role Decision: "Follower." Motion Instruction: "Reduce speed to 3.0 m/s to maintain safe distance."
	
	\textbf{Example 2:}
	\faMap: "No vehicles ahead. $CAV_3$ should accelerate to merge smoothly."
	\faTags: Role Decision: "Leader." Motion Instruction: "Accelerate to 5.0 m/s and maintain lane position."
	
	\faLightbulbO: "Based on the current scenario, assign roles and propose motion instructions for all vehicles." \end{tcolorbox}

\textbf{2. Contextual Reasoning:} Context is crucial for tailoring responses to realistic settings. Weather, time of day, and road congestion are embedded in prompts to enable nuanced planning.

\begin{tcolorbox}[colback=gray!5!white, colframe=gray!75!green, title=\textbf{Example with Contextual Cues for Traffic Coordination in Merging Scenarios}] \centering \faMap: Scenario Context \hfill \faTags: Role and Instructions \hfill \faLightbulbO: Prompt \hfill \faCar: LLM Response
	\raggedright \rule{13cm}{1pt} \linespread{1.25} \selectfont
	\small
	\faMap: "The current traffic merging scenario occurs during evening rush hour under rainy weather conditions. Vehicles are entering from an on-ramp with heavy congestion on the main road. Vehicle $CAV_1$ is at 11.75 m, moving at 3.5 m/s. Vehicle $CAV_2$ is behind at 8.0 m."
	
	\faTags: Assign roles and propose motion strategies to minimize delays and avoid collisions.
	
	\faLightbulbO: "Based on the scenario, assign roles (Leader, Follower) to the vehicles and provide motion instructions to ensure smooth merging flow while accounting for rain-induced low traction."
	
	\faCar: $CAV_1$: Role: Leader. Instruction: "Maintain current speed of 3.5 m/s to merge into the main road safely and avoid abrupt braking."
	$CAV_2$: Role: Follower. Instruction: "Reduce speed to 3.0 m/s and maintain a safe 5-meter headway to prevent rear-end collisions due to low traction."
	
\end{tcolorbox}

\textbf{3. Integration with CoMAL Framework:} In CoMAL, a \textit{Memory Module} stores previously observed traffic strategies, while a \textit{Collaboration Module} enables vehicles to share and refine their decisions dynamically. The few-shot examples embedded in prompts are augmented by shared agent observations, ensuring the LLM has context-aware input to reason effectively in diverse traffic scenarios.
FSL prompts are enhanced by this shared context, enabling the LLM to plan cooperatively across agents.

\begin{figure}[h!]
	\centering
	\includegraphics[width=0.99\linewidth]{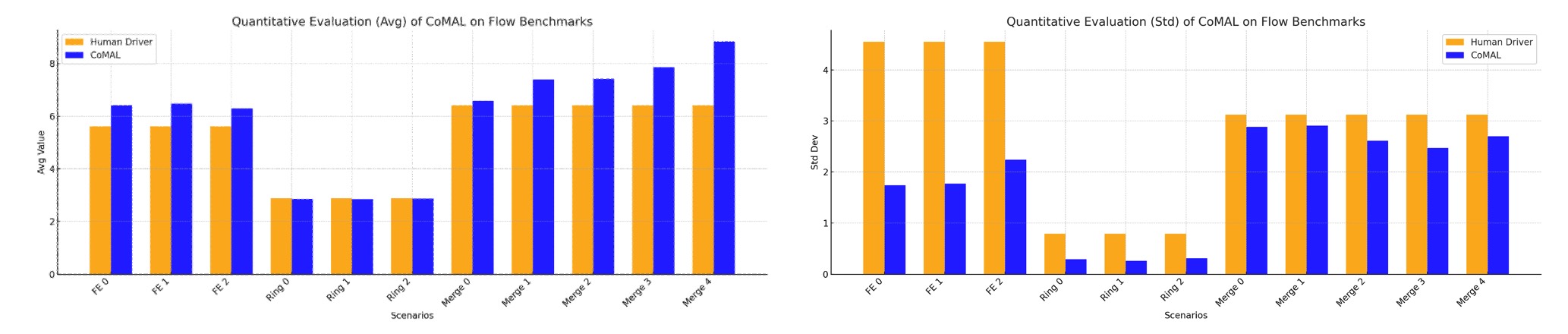}
	\caption{The comparison between CoMAL with human drivers, from the experiment, we could observe that the cooperative multi-agent LLM is performing consistently better than human drivers in the Flow benchmark environment~\cite{vinitsky2018benchmarks}.}
	\label{fig:comal_compare}
\end{figure}

\paragraph{Demonstration and Results} Experiments in CoMAL show that LLMs guided by few-shot examples outperform intuitive strategies and achieve smoother traffic flow. As an example in Figure.~\ref{fig:comal_compare}, in the various Merge, Ring, Figure Eight (FE) Scenarios, the system consistently demonstrated improved average vehicle speeds and reduced speed variance compared to human drivers, and even reinforcement learning baselines as discussed in~\cite{yao2024comal}.

\paragraph{Comparison with Traditional Methods}
Traditional approaches to traffic coordination often rely on rule-based logic or control-theoretic models. In contrast, FSL with LLMs offers rapid deployment without retraining, adaptability to unseen environments, and seamless integration with real-time, context-rich decision-making.

\paragraph{Strengths and Challenges}
FSL supports real-time planning in dynamic traffic, scales to new scenarios with a few examples, and requires no additional labeled datasets. \newline
Dispite its strengths, its performance depends on prompt quality, failure cases can be difficult to interpret or debug, and it needs careful evaluation in safety-critical contexts.

Few-shot in-context learning enables generative AI to coordinate connected autonomous vehicles in complex traffic environments. Through carefully designed prompts and context embedding, LLMs can generate cooperative strategies for traffic merging and other coordination tasks. The approach provides scalable, adaptive, and interpretable decision-making support in mixed-autonomy traffic systems.

\subsection{Other Techniques Enhancing LLM Inference}~\label{other-inference}

\noindent \textbf{Retrieval-Augmented Generation (RAG)}.
Retrieval-Augmented Generation (RAG) enhances the accuracy and relevance of generative AI outputs by dynamically integrating external data sources during inference~\cite{wang2024evaluating, qi2024model}. Unlike static models that rely solely on pre-trained knowledge, RAG combines real-time retrieval of information from external databases with text generation, ensuring the outputs remain contextually accurate and up-to-date. In transportation planning, RAG can be particularly valuable for dynamic tasks such as real-time traffic analysis or policy impact assessments~\cite{wu2024medical, peng2024graph}.

For example, when tasked with optimizing traffic signal timings during high-demand periods, a RAG-enhanced generative model can retrieve live traffic counts, historical congestion trends, and environmental factors (e.g., weather conditions). The retrieved information is incorporated into the model’s input, allowing it to generate traffic signal adjustments that are grounded in current data and actionable in real-world scenarios. This approach ensures generative AI systems adapt to rapidly evolving conditions in transportation networks while maintaining precision and relevance.

\vspace{1mm}
\noindent \textbf{Chain-of-Thought Reasoning}
\label{COT}
Chain-of-Thought (CoT) reasoning improves generative AI's ability to handle multi-step, complex transportation optimization problems by breaking them into sequential reasoning steps~\cite{zhanggenerating}. This technique reduces oversimplification and improves the transparency of AI outputs, making it particularly suitable for decision-making in transportation systems.

\vspace{1mm}
\noindent \textbf{Knowledge Editing.} Knowledge Editing allows for dynamic updates to the internal representations of a generative AI model without retraining the entire model~\cite{zhang2024oneedit}. In transportation applications, where real-time data and policies frequently change, Knowledge Editing enables targeted updates to specific components of the model’s knowledge base.

For example, consider a scenario where a city introduces a new congestion pricing policy in its central district. Instead of retraining the entire model, Knowledge Editing can inject the updated policy details directly into the model's representation of city traffic rules. When tasked with generating congestion mitigation plans, the model now incorporates the new pricing constraints into its outputs, ensuring recommendations align with the latest policies. This flexibility allows generative AI systems to remain responsive to real-time updates, such as changes in infrastructure, traffic regulations, or environmental policies.

\vspace{1mm}
\noindent \textbf{Self-Consistency Decoding.} Self-Consistency Decoding improves the robustness and reliability of generative AI outputs by prompting the model multiple times with slightly varied initial conditions, generating diverse candidate responses, and selecting the most consistent answer~\cite{huang2023enhancing}. This method reduces the randomness often inherent in generative models, particularly in tasks involving optimization or decision-making.

In transportation applications, such as intersection signal optimization, Self-Consistency Decoding can stabilize recommendations. For instance, a generative model tasked with adjusting traffic light cycles under heavy congestion may initially produce varying outputs depending on the prompt phrasing. Self-Consistency Decoding generates multiple solutions and selects the one most frequently proposed or aligned with traffic optimization criteria. This approach ensures that the final output reflects a robust, consensus-driven solution that minimizes delays and improves traffic.

\begin{figure*}[h!]
	\centering
	\includegraphics[width=1\linewidth]{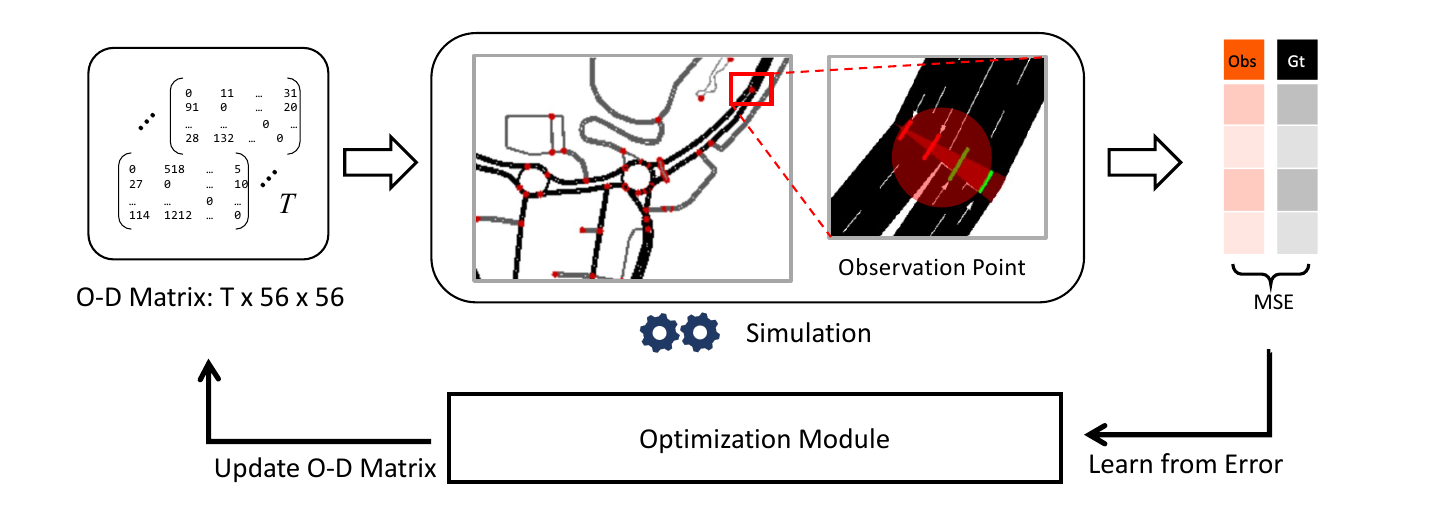}
	\caption{The overview of the OD matrix task in this case study.}
	\label{fig:problems}
\end{figure*}

\subsection{Case Study: LLM's Ability in OD-Calibration in Terms of Feature Quality}
\label{caseStudy}

This subsection presents a case study on how LLMs perform in Origin-Destination (OD) Calibration tasks for transportation planning. Specifically, we utilize an OD calibration dataset from Open Traffic Intelligence (OpenTI)~\cite{da2024open}, which includes initial OD matrices and observed traffic counts, to evaluate how generative AI models perform when tasked with OD calibration. The study focuses on two critical aspects: (1) Exploring the few-shot learning solution designs in current LLMs for solving the OD calibration task, and (2) assessing the quality of model-generated features (observation counts) compared to ground-truth OD matrix's count feature, an overview of the problem is shown in Figure~\ref{fig:problems}.

\subsubsection{Task Description}
In this section, we consider an origin-destination (OD) matrix $\mathbf{X}$ of dimension $T \times 56 \times 56$, where $T$ is the total number of discrete time slices, and $56$ represents the number of blocks in the city. Specifically, $\mathbf{X}$ contains the entries $x_{t,i,j}$ for $t = 1, \dots, T$ and $i,j \in \{1,\dots,56\}$, where $x_{t,i,j}\ge0$ denotes the traffic demand (number of vehicles) from block $i$ to block $j$ during time slice $t$. Once the full matrix $\mathbf{X}$ is specified, a traffic simulator~\cite{zhou2014dtalite} is executed to generate network flows; however, since the observation point is sparse, the effect from the $f\bigl(\mathbf{X}\bigr)$ is very limited, which brings much difficulty to this task. We denote $O_t$ as the ground-truth measurement at the sensor for time slice $t$. Our goal is to calibrate the unknown entries of $\mathbf{X}$ so as to minimize the mean squared error between the simulator-derived flows and the real observations. Concretely, we define the following: 

\begin{equation}\label{eq:od}
	\min_{\{x_{t,i,j}\}}
	\frac{1}{T}\sum_{t=1}^T 
	\bigl(
	O_t - f\bigl(x_{t,\cdot,\cdot}\bigr)
	\bigr)^2,
	\quad
	\text{subject to}
	\quad
	x_{t,i,j} \ge 0.
\end{equation}

where, $f\bigl(x_{t,\cdot,\cdot}\bigr)$ denotes the effective flow from simulated traffic that makes differences to observations at the sensor location corresponding to the OD submatrix (time-slice based OD) $\{x_{t,i,j}\}_{i,j=1}^{56}$. The discrepancy between the observed and simulated flows is therefore measured by the \emph{mean squared error} (MSE) at each time step, and the average of these errors over all $T$ time slices forms the objective function in Eq.~\ref{eq:od}.

\subsubsection{Model Configurations, Computational Resources, and Dataset Selection}

We conduct exploration on the open-sourced and widely-used generative AI models \textbf{Llama 3.1-70B}~\cite{singh2024scidqa}—to simulate OD calibration scenarios. Experiments are conducted on hardware configurations tailored for large-scale traffic data tasks and are deployed on a server with 8 NVIDIA A100 GPUs, 2 AMD EPYC CPUs, and 2 TB memory.

The dataset used in this case study is from \textbf{OpenTI}~\cite{da2024open}, a benchmark traffic dataset featuring initial OD matrices, observed traffic flows, and can generate calibrated OD outputs for varying network conditions. This dataset captures dynamic traffic patterns across different time intervals, making it ideal for evaluating biases and feature quality in generative OD calibration tasks.

\subsubsection{Experimental Design and Methodology}

The case study employs two sets of experimental settings, one is the traditional method (Genetic Algorithm), and the other one includes two LLM-based methods. 
\begin{table}[t!]
	\centering
	\caption{Genetic Algorithm (NSGA2) Hyperparameters for OD Calibration}
	\begin{tabular}{cc}
		\toprule
		\textbf{Parameter} & \textbf{Value} \\
		\midrule
		Number of Iterations & 300 \\
		Population Size & 600 \\
		Mutation Eta & 0.1 \\
		Crossover Eta & 0.6 \\
		Crossover Probability & 0.9 \\
		\bottomrule
	\end{tabular}
	
	\label{tab:paramGenetic}
\end{table}

\textbf{Method Type 1: Genetic Algorithm for OD Calibration.} 
To serve as a baseline solution for the OD calibration problem, we employ a genetic algorithm (GA) using the NSGA2~\cite{blank2020pymoo}. The objective is to minimize the mean squared error (MSE) defined in Eq.~\ref{eq:od}, where each candidate solution represents an estimated OD matrix \(\mathbf{X}\). Given the large search space and non-linear mapping from \(\mathbf{X}\) to the observed sensor measurements \(O_t\), NSGA2 serves as an efficient optimization strategy to iteratively refine OD matrices that better fit the observed data.

\begin{figure*}[t!]
	\centering
	\includegraphics[width=0.79\linewidth]{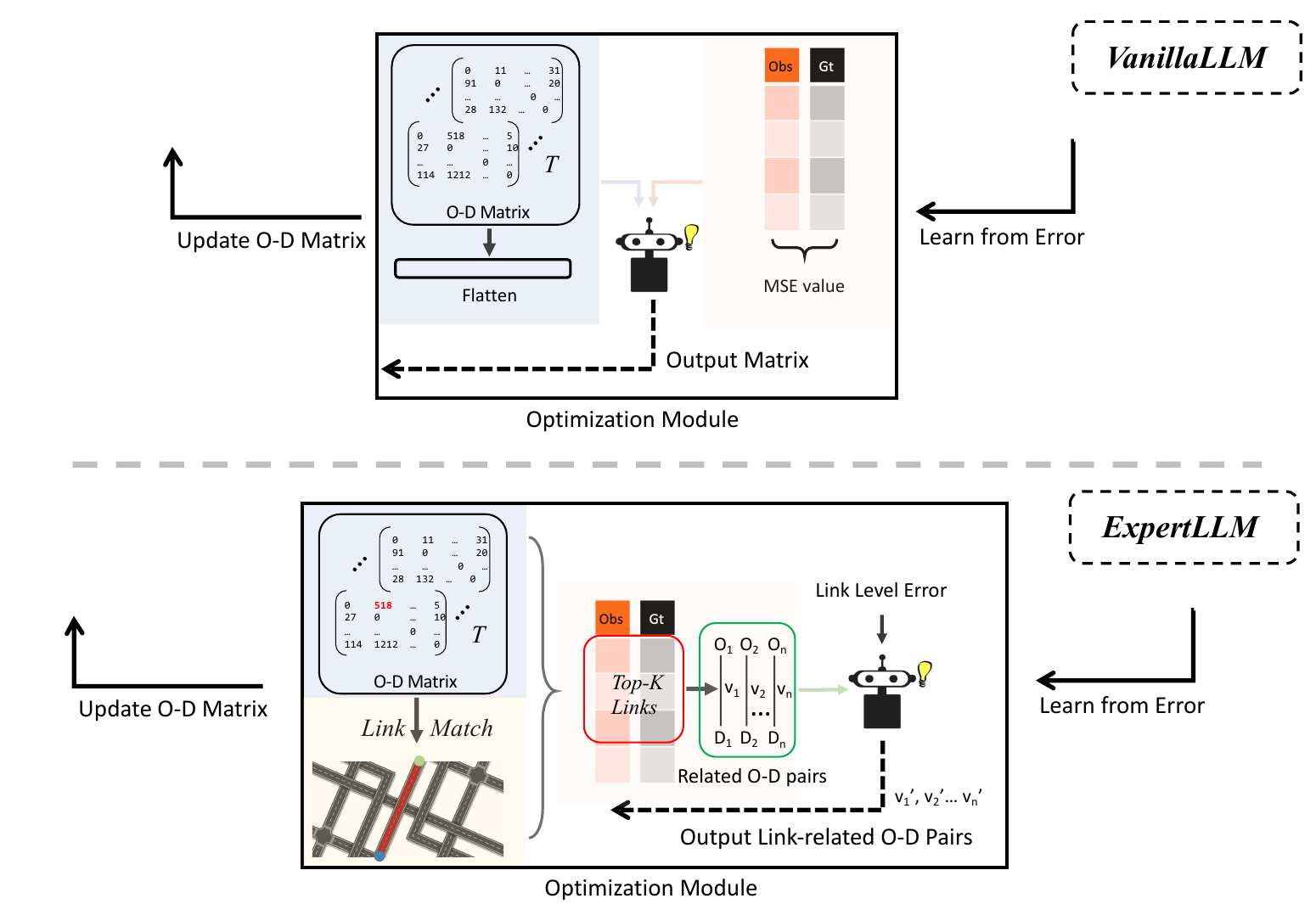}
	\caption{The difference between the methodology structure of \textbf{VanillaLLM} and \textbf{ExpertLLM}. As shown in the image, the VanillaLLM directly inputs the flattened OD matrix to the LLM, together with the MSE value, it provides a vast space to optimize given the dimension of 56 x 56 = 3136 for a single time step, and each value is not bounded, this is a very challenging task for LLM to reason. But ExpertLLM first builds the connection from the OD matrix to link performance and uses human expert's domain knowledge on the top-k links searching (which reduces the observation space), then it tracks back to the OD pairs that contribute to these k links, and LLM is asked to solve a problem on a link basis, which is a much easier task for LLM to infer the adjustment direction, this relaxed the difficulty of the proposed question by leveraging necessary information and combine the analysis with reasoning of LLMs.}
	\label{fig:prompts}
\end{figure*}

Each candidate OD matrix \(\mathbf{X}\) is represented as a chromosome \(\boldsymbol{\chi} \in \mathbb{R}^{T \times 56 \times 56}\), encoding all \(x_{t,i,j}\) values in a flattened vector form. The genetic algorithm initializes a population of such candidate solutions and iteratively evolves them using three key operations: selection, crossover, and mutation. The fitness of each candidate solution is determined by evaluating the simulated sensor flows \(f\bigl(x_{t,\cdot,\cdot}\bigr)\) and computing the corresponding MSE following the Eq.~\ref{eq:od}. Formally, the GA optimization in our problem follows these steps:
(1)~\underline{Initialization}: A population of \(N\) random OD matrices \(\mathbf{X}^{(k)}\), where \(k = 1, \dots, N\), is generated within predefined bounds (e.g., \(x_{t,i,j} \geq 0\)).
(2)~\underline{Fitness Evaluation}: Each OD matrix \(\mathbf{X}^{(k)}\) is fed into a traffic simulator to generate predicted flows \(\widehat{O}_t^{(k)} = f\bigl(x_{t,\cdot,\cdot}^{(k)}\bigr)\). The fitness of each solution is computed by Eq.~\ref{eq:od}.
(3)~\underline{Selection}: Parents are selected based on their fitness scores using NSGA2’s non-dominated sorting and crowding distance mechanisms. Solutions with lower MSE values are more likely to be chosen.
(4)~\underline{Crossover and Mutation}: New offspring solutions are generated by recombining parent chromosomes through crossover. Additionally, a mutation operator perturbs selected elements of \(\boldsymbol{\chi}\) to introduce diversity: $
x_{t,i,j}^{\text{new}} = x_{t,i,j} + \epsilon, \quad \epsilon \sim \mathcal{N}(0, \sigma^2)$, 
where \(\sigma\) controls the mutation strength.
(5)~\underline{Survivor Selection}: The offspring solutions replace the least fit individuals in the population, ensuring that better solutions persist in subsequent generations.

This iterative process continues until a predefined convergence criterion is met (we set it based on the Interaction steps = 250), with details of setting in Table.~\ref{tab:paramGenetic}. The final solution \(\mathbf{X}^*\) represents the best-calibrated OD matrix that minimizes the discrepancy between simulated and observed traffic flows.

\textbf{Method Type2: LLM-based OD Calibration Methods.} 
In this section, we have two different designs of LLM-based methods as an exploration. The first one, named \textbf{VanillaLLM}, is the most straightforward design that, we directly pass the current solution with shape $(T \times 56 \times 56)$ and the observed link performance, as well as the calculated gap, and the prompt design is as shown in prompt \ding{172}. The second design is based on the combination of domain expert knowledge and LLM reasoning, which tries to simplify the problem by leveraging heuristic information to reduce the search space and encouraging LLM's effective reasoning, named \textbf{ExpertLLM}. The difference between the two LLM-related solvers' structures is as shown in Figure~\ref{fig:prompts}.

\noindent The following example prompts illustrate the CoT-based OD calibration design:

\begin{tcolorbox}[colback=gray!5!white, colframe=gray!75!green, title=\textbf{\ding{172} Prompt for VanillaLLM solving OD Calibration}]
	\small
	\raggedright
	\textcolor{boxblue}{\textbf{System Description:}}  
	We are calibrating a 56×56 Origin-Destination (OD) matrix for a transportation network. Each entry [i, j] represents trips from origin i to destination j.
	
	\textcolor{boxblue}{\textbf{Known Information:}}
	
	1. Current OD matrix (56 lines): \{matrix text\}
	
	2. Current MSE: \{current MSE\}
	
	\textcolor{boxblue}{\textbf{Logics:}}
	
	A lower MSE indicates that the OD matrix better reflects real-world traffic patterns.
	
	\textcolor{boxblue}{\textbf{Task Description:}}
	
	Focus on analyzing patterns and identifying actionable changes to improve matrix accuracy and reduce MSE.  
	
	\textcolor{boxblue}{\textbf{Return Constraints:}}  
	Only reply with 56 lines of a new 56×56 matrix.

	\textcolor{boxblue}{\textbf{Example:}} Adjusted OD Matrix (for 2x2 shape): $ \left[ \begin{smallmatrix} 320 & 180 \\ 160 & 340 \end{smallmatrix} \right] $.

	\textbf{Output:} Final \textit{\textbf{OD matrix}} with error reduction.
\end{tcolorbox}

\subsubsection{Results and Analysis}

\begin{figure}
	\centering
	\includegraphics[width=0.9\linewidth]{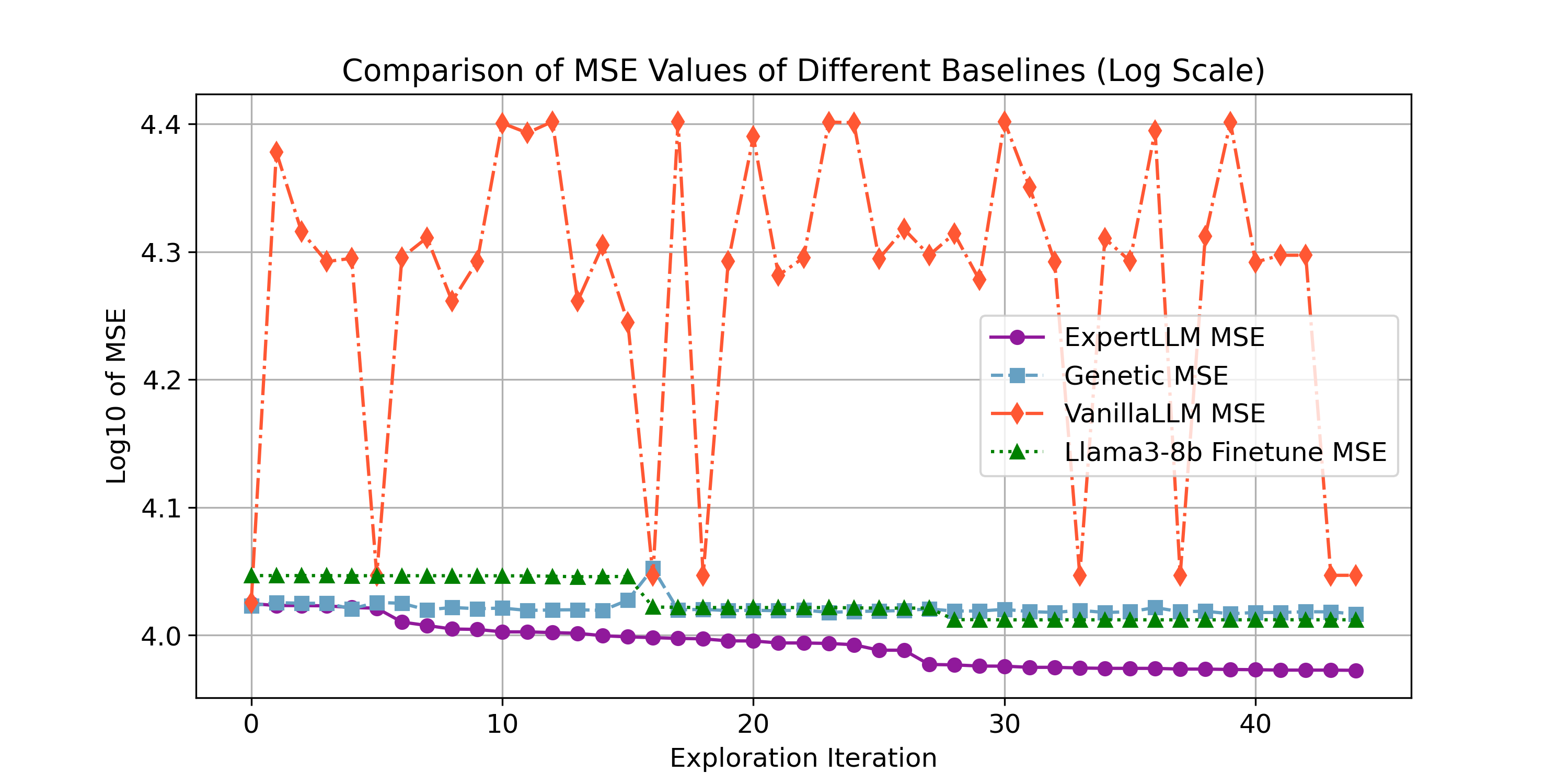}
	\caption{The performance over iterations on three methods, we could observe that, given the same amount of the iterations, the \textbf{ExpertLLM} is outperforming the genetic algorithm and the \textbf{VanillaLLM}. Another performance is from the fine-tuning techniques of LLM on Llama3-8b (\textbf{Llama3-8b Finetune MSE}), we can see that, the smaller model (8 billion parameters model) can achieve a performance of genetic algorithm and even better than some of the larger model's ability such as VanillaLLM.}
	\label{fig:compareLLMs}
\end{figure}

% \noindent \textbf{Calibration Bias in OD Matrices.}  
As shown in Figure.~\ref{fig:compareLLMs}, it compares the mean square error (MSE) for three different baselines of OD matrix calibration under the same iterations. 

- \textbf{For the Genetic Algorithm-based optimization module}, it made a stable improvement with a slow evolving process (we only show the first 50 iterations for a fair comparison), we have verified that, this algorithm can eventually lead to a relatively good performance after more than 100,000 populations given a test time of roughly a week, this method is not efficient due to the large exploration space and vast population-related operations, such as cross-over and mutate. In addition, extensive experiments revealed that while the algorithm steadily improves the solution quality over many generations, its convergence is exceptionally slow and the overall computational cost remains high. The algorithm relies on repeated cycles of selection, crossover, and mutation, each of which introduces randomness that both aids exploration and hinders rapid convergence. As a result, despite achieving promising results after a prolonged period, the method demands a significant amount of computational resources and time—often taking nearly a week to complete a comprehensive search across the vast parameter space. Furthermore, the performance is highly sensitive to parameter tuning, and even minor adjustments in mutation rates or crossover probabilities can lead to substantial variations in outcomes. This inherent inefficiency, driven by the enormous search space and the complexity of managing large populations, underscores the necessity for more efficient hybrid approaches that could combine global exploration with faster local optimization techniques.

\begin{tcolorbox}[colback=gray!5!white, colframe=gray!75!green, title=\textbf{\ding{173} Prompt for ExpertLLM solving OD Calibration}]
	\small
	\raggedright
	\textcolor{boxblue}{\textbf{System Description:}}  
	We are calibrating a 56x56 Origin-Destination (OD) matrix for a transportation network. Each entry [i, j] represents the number of trips from origin i to destination j. This OD matrix is used in a simulation to generate traffic volume counts on various links in the network.
	
	You will be provided with the following details of one link:
	
	- Link ID: {link id}
	
	- The simulated volume: {simulated vol}
	
	- The ground truth volume: {obs count}
	
	- The absolute error, which is calculated as: abs(Simulated Volume - Ground Truth Volume) = {abs error}

	\textcolor{boxblue}{\textbf{Known Information:}}
	
	The following \{sample size\} randomly sampled OD elements (i, j) that contribute to this link, along with their current flow values: \{pairs str\}
	
	\textcolor{boxblue}{\textbf{Logics:}}
	
	A lower MSE indicates that the OD matrix better reflects real-world traffic patterns.
	
	\textcolor{boxblue}{\textbf{Task Description:}}
	
	Adjust ONLY these OD elements' flow values to reduce the absolute error, thereby improving the alignment between the simulation results and real-world traffic observations.

	\textcolor{boxblue}{\textbf{Return Constraints:}}  
	
	- Do not return any placeholder text.
	
	- Return ONLY the updated values of the {sample size} OD elements with their indices, one per line, in the format:
	
	[(i, j), new value]
	
	[(i2, j2), new value 2]

	\textbf{Output:} Updated \textit{\textbf{OD values for selected links}}.
\end{tcolorbox}

- \textbf{For VanillaLLM},  it clearly struggled to optimize the OD Matrix, with MSE values ranging from around 94,000 to well over one million. Despite having full access to the entire OD Matrix and its corresponding MSE, it was unable to derive the actionable insights needed for effective optimization. Intriguingly, the lowest MSE occurred when VanillaLLM generated an all-zero matrix—contrary to explicit instructions—demonstrating its limited capacity to adhere to constraints and reason effectively. This outcome strongly highlights the necessity for human expert intervention to guide and refine the optimization process. Furthermore, our experiments revealed that although VanillaLLM can process extensive data, its internal mechanisms are not well-equipped to handle the complex structure of the OD Matrix, as the model repeatedly defaulted to trivial solutions that minimized error numerically but failed to capture any meaningful spatial or operational nuances. In several cases, the algorithm converged prematurely to an all-zero output, a clear sign of its inability to balance multiple constraints and the multidimensional interplay of variables, resulting in erratic behavior with unpredictable oscillations between high error values and degenerate solutions. This inconsistency suggests that VanillaLLM is overwhelmed by the high dimensionality and vast exploration space inherent in the task, ultimately necessitating a more guided, hybrid approach that incorporates domain-specific heuristics and human oversight to steer the optimization toward robust, contextually aware solutions.

- \textbf{For ExpertLLM}, as shown in Figure.~\ref{fig:compareLLMs}, it achieved significant MSE reductions by focusing on the top K links with the highest absolute error, a strategy that underscores its ability to selectively refine the most critical components of the OD Matrix. This targeted approach not only improves performance by concentrating computational resources on the most problematic links but also demonstrates the model’s potential to deliver rapid improvements in key areas where errors are most obvious. However, despite these promising gains, ExpertLLM eventually reaches a point of diminishing returns where further reductions in MSE plateau, indicating that the automated process alone cannot fully capture the nuanced interdependencies present in the data. At this stage, the benefits of continued unsupervised optimization wane, and the system’s performance becomes constrained by its inability to adapt its focus beyond the initially selected top-K links. In such cases, human intervention becomes crucial to steer the LLM toward more promising avenues for improvement. For instance, replacing the automated top-K selection with input from a transportation planning expert could enable the LLM to identify and prioritize not only the most error-prone links but also those that are strategically important within the overall network, ensuring that improvements have a broader impact on system performance. Moreover, expert guidance might help in reweighting or redefining the criteria for link selection based on real-world constraints and operational priorities, thereby fostering a more context-aware and adaptive optimization process. By integrating domain expertise with automated algorithms, the hybrid approach promises to overcome the plateau in performance, ultimately leading to a more robust and effective optimization of the OD Matrix.

\subsubsection{Summary and Insights}
Overall, ExpertLLM demonstrates a significant improvement over both VanillaLLM and the 
Genetic Algorithm in optimizing the OD Matrix. While the ExpertLLM pipeline shows 
considerable potential, the performance of both systems ultimately underscores the importance of 
incorporating human expertise—particularly when improvements plateau—to achieve robust and 
scalable optimization. Integrating human insight not only enhances the interpretability of LLM 
suggestions but also provides the necessary domain-specific adjustments to drive further 
reductions in MSE.

\begin{figure}
	\centering
	\includegraphics[width=0.9\linewidth]{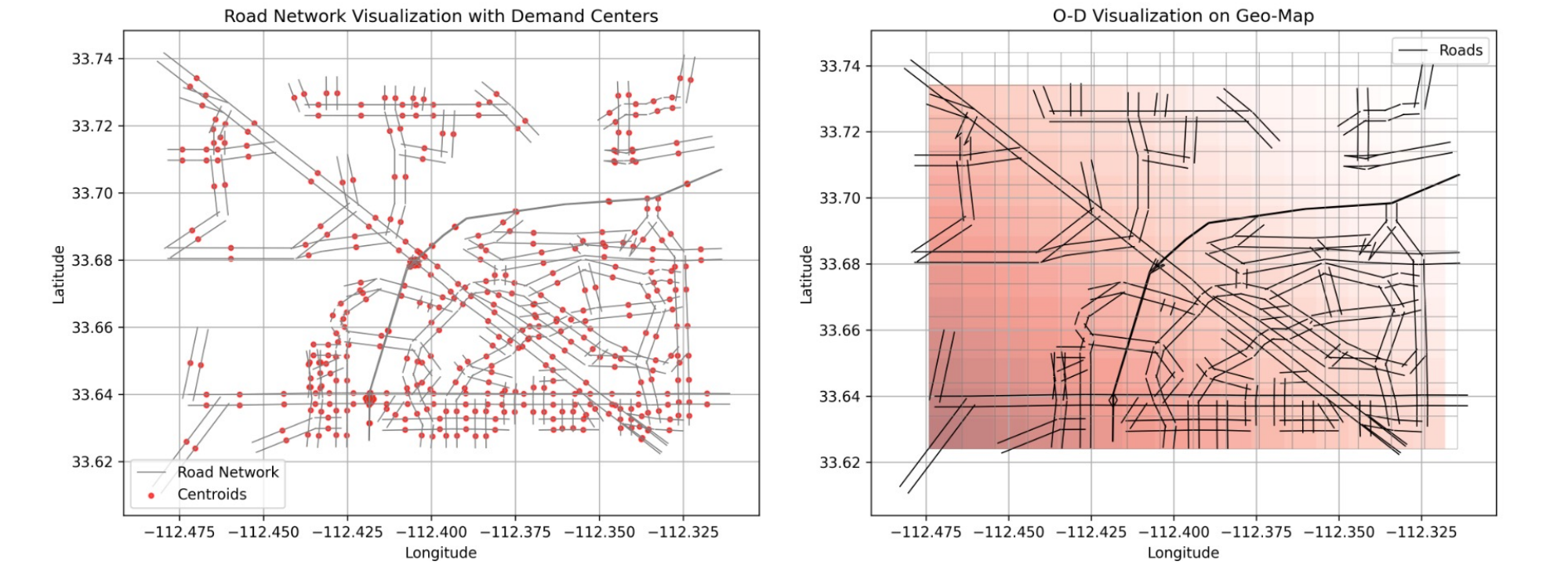}
	\caption{The visualization of a basic geographic map (left) with the OD heatmap (right), as shown on the right hand, which is an example with mock data showing that the darker color it is, the more traffic demand in the area. While effective in showing the overall traffic activities, it can hardly show the direction of trip movements, so we employ the following Figure~\ref{fig:OD2} for more detailed and realistic visualization.}
	\label{fig:OD}
\end{figure}

\begin{figure}
	\centering
	\includegraphics[width=0.99\linewidth]{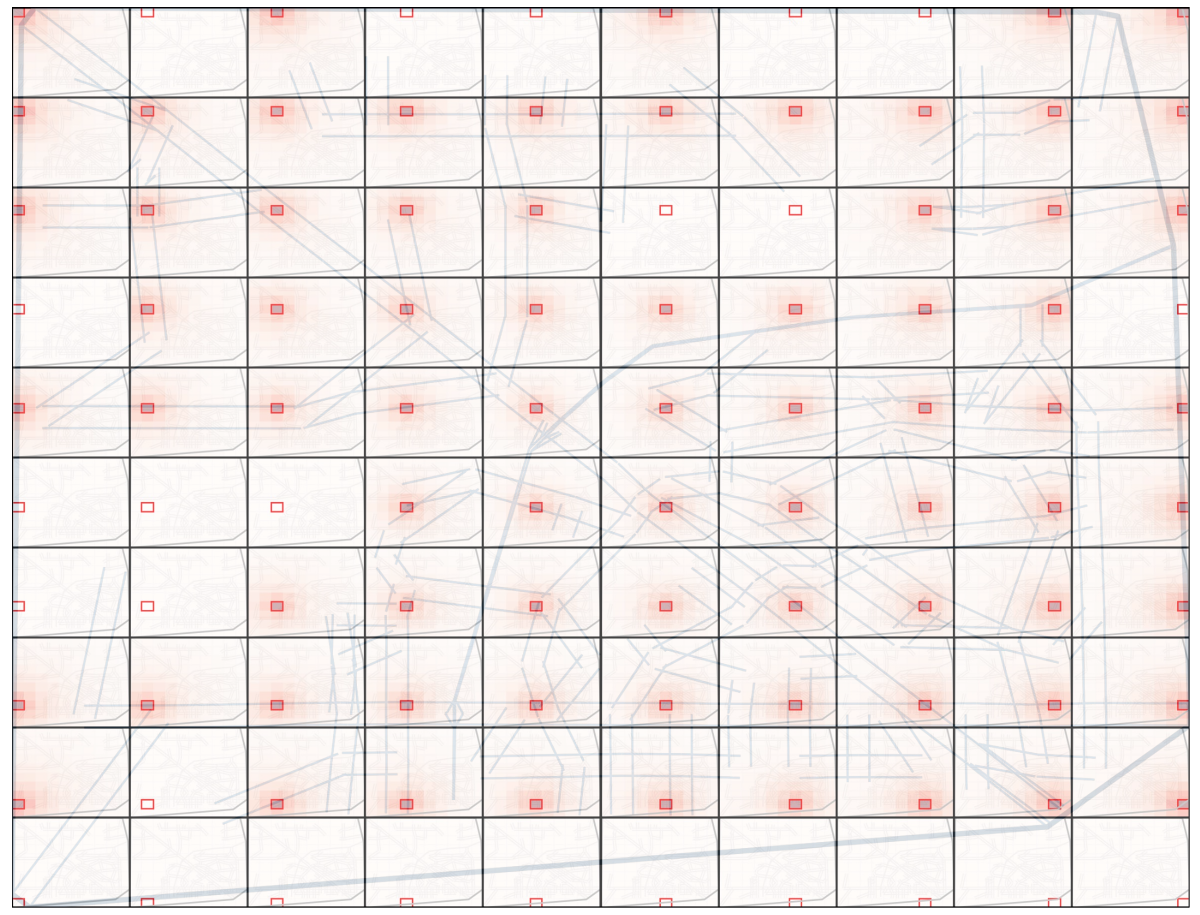}
	\caption{The visualization of the OD heatmap with basic geographic map, the darker color shows the higher demand in the OD matrix. The entire figure shows the map of the studied location, and it is splitted into 10 times 10 sub areas, in each area, it contains a complete map of entire area as well, with highlighted area as the demand from this area to this same local area, which normally the highest, and the lighter color in neighbor cells show the demand travel from the highlight area to others~\cite{wood2010visualisation}.}
	\label{fig:OD2}
\end{figure}

This case study highlights the strengths and limitations of generative AI models in OD calibration tasks:  

- \textbf{Strengths}: 1. The larger the LLMs (model parameters), the better the performance it generally demonstrates during the OD calibration tasks. This is mainly revealed in two dimensions: better instruction-following abilities and longer input/output token processing abilities. 2. The LLMs can be easily prompted and fit human-involved designs and bring many performance improvements. 3. Even though the larger model requires expensive computing resources, the smaller model can be fine-tuned for outstanding performance.

- \textbf{Limitations}: 1. The LLMs take a longer time in the reasoning process, especially when the input/output string is long or contains special requirements regarding the format. 2. The training-related operations require specific efforts to learn efficient training strategies, such as LoRA fine-tunning~\cite{li2023loftq}. 3. During fine-tuning, privacy information could be released in terms of the data use~\cite{chen2024privacy}.

The results emphasize the importance of integrating intermediate feature generation steps, such as those enabled by Chain-of-Thought reasoning, to enhance OD calibration. Future work will explore methods to further mitigate planning biases and improve model generalizability across diverse traffic networks.

\section{Future Directions \& Challenges} \label{sec:challenges}

\subsection{Possible Pitfalls of Generative AI for Transportation Planning}\label{pitfalls} While the potential benefits of Generative AI in transportation planning are substantial, its integration raises several challenges. Ensuring data quality, addressing biases in model outputs, and achieving model interpretability are critical hurdles that must be addressed to ensure reliable and equitable outcomes. One particularly daunting challenge is building the comprehensive databases from which Generative AI models learn. This includes tasks such as aggregating data from multiple jurisdictions, reconciling inconsistencies across datasets, and performing extensive labeling, as discussed in Section~\ref{dataset-preparation}. The labor-intensive nature of these processes, combined with the need for high granularity and accuracy, often limits the scalability of AI applications in transportation.

Furthermore, the scalability of these tools must be carefully considered to ensure their applicability across diverse transportation contexts, ranging from urban traffic management to interstate infrastructure planning. Scalability is especially challenging in cases where localized variations in traffic patterns, infrastructure constraints, and population behaviors require fine-tuned adaptations of generative models.

\begin{figure}[t!]
	\centering
	\vspace{-5mm}
	\includegraphics[width=0.98\textwidth]{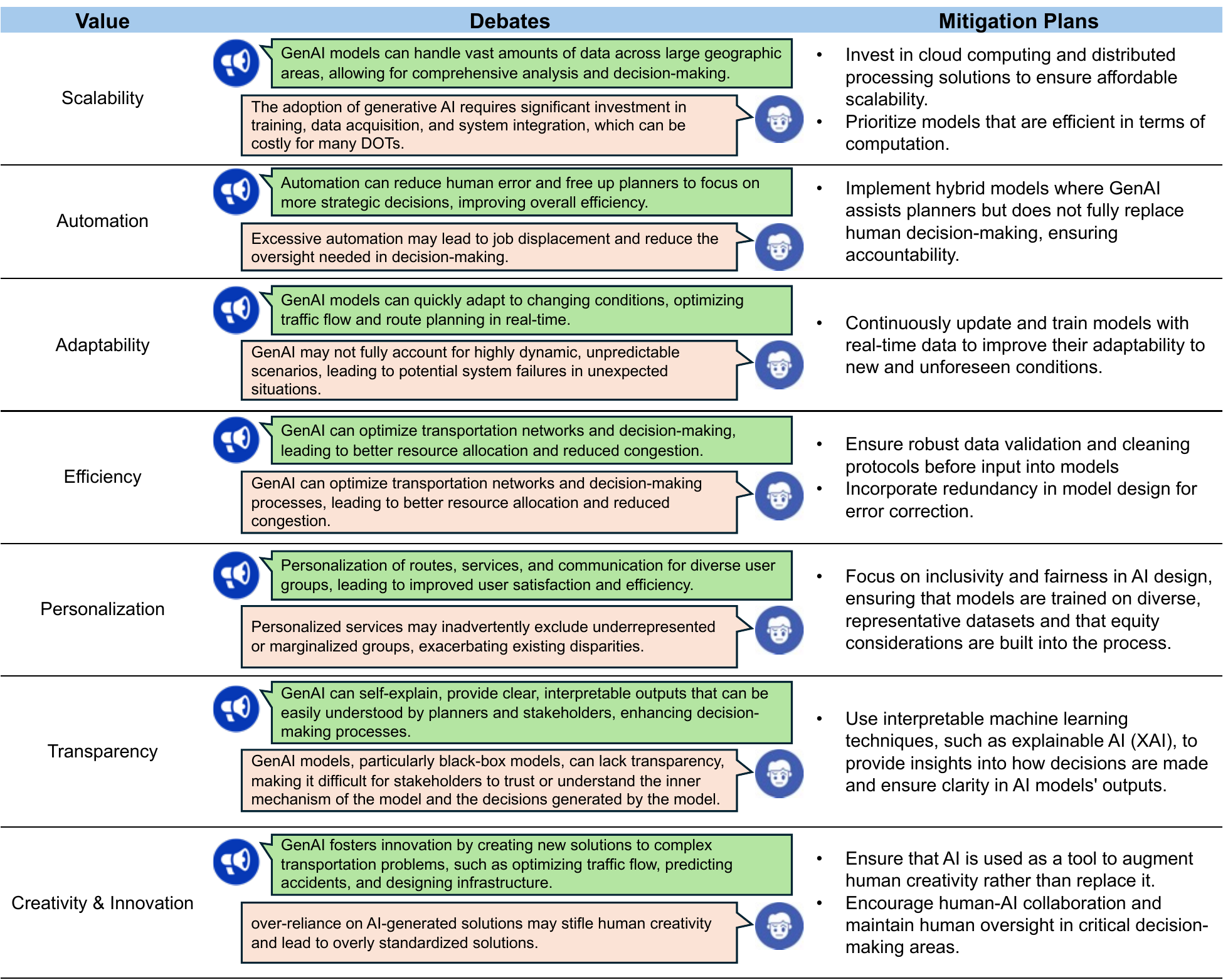}
	\caption{Debates over the potentials of GenAI in transportation planning and their mitigations}
	\label{fig:genai-tech-trans}
\end{figure}

Figure~\ref{fig:genai-tech-trans} adapts the key value propositions of Generative AI, emphasizing its critical features tailored to the unique demands and values of transportation planning. These features highlight the need for innovative strategies to address the aforementioned pitfalls while fully leveraging the transformative potential of Generative AI in transportation systems.

\subsection{Pipelines of Integrating Transportation Planning with Generative AI}\label{pipelines}

The integration of generative AI into transportation planning offers transformative opportunities but presents challenges that stem from the domain's complexity. Generative models excel at automating tasks such as \textit{traffic demand calibration}, \textit{policy impact evaluation}, and \textit{multi-modal optimization}. However, fully adapting these models to the spatial, temporal, and regulatory requirements of transportation systems remains a major hurdle~\cite{yao2024comal,da2024open}.

A key challenge arises from the limited ability of general-purpose generative AI models to incorporate domain-specific constructs, such as traffic flow dependencies, vehicle heterogeneity, and infrastructure constraints. Future advancements must focus on domain-aware fine-tuning and hybrid pipelines that integrate generative AI models with traditional transportation methods, ensuring that outputs are reliable, interpretable, and actionable.

\noindent \textbf{Modularized Pipelines for Transportation Planning.}  
Modularized pipelines offer a systematic framework to handle transportation tasks by decomposing complex problems into smaller, manageable components. Each module is optimized for specific objectives, improving scalability, interpretability, and accuracy. For example, a \textit{real-time traffic optimization pipeline} can be structured as follows:

\begin{enumerate}[leftmargin=*]
	\item \textbf{Data Preprocessing Module:} Integrates data from sensors, GPS logs, weather feeds, and socio-economic reports. Preprocessing includes data cleaning, normalization, and spatial alignment.  
	\item \textbf{Feature Generation Module:} Uses generative AI to predict traffic demand, accident probabilities, or multi-modal travel behaviors based on real-time inputs.  
	\item \textbf{Optimization Module:} Leverages generative AI outputs for optimizing traffic control strategies, such as adaptive signal timing or congestion rerouting.  
	\item \textbf{Evaluation Module:} Assesses the outcomes using performance metrics like travel time reduction, fuel savings, and network throughput.  
\end{enumerate}

These modularized pipelines facilitate end-to-end solutions for tasks like congestion prediction, OD calibration, and network optimization, enabling planners to customize workflows for region-specific transportation needs.

\noindent \textbf{Agentic RAG-Enabled Pipelines.}  
Integrating \textit{Agentic Retrieval-Augmented Generation (Agentic RAG)}~\cite{lewis2020retrieval} into transportation pipelines enhances generative AI’s ability to incorporate real-time and domain-specific knowledge dynamically~\cite{wang2024evaluating}. RAG-based frameworks retrieve contextual information—such as traffic incidents, weather disruptions, or policy updates—from external databases, ensuring that generative AI outputs are timely and grounded in accurate data. 

For instance, a RAG-enabled pipeline for \textit{infrastructure resilience planning} can include:
\begin{enumerate}[leftmargin=*]
	\item \textbf{Retriever Agent:} Dynamically retrieves external knowledge, such as live incident reports, historical disruptions, and infrastructure performance metrics.  
	\item \textbf{Integration Module:} Combines retrieved knowledge with generative prompts to generate adaptive infrastructure strategies.  
	\item \textbf{Resilience Assessment:} Simulates system responses under various scenarios (e.g., floods, earthquakes) and provides recommendations for infrastructure upgrades or policy changes.  
\end{enumerate}

Agentic RAG-based pipelines can dynamically adapt to new information, improving model accuracy and decision-making in time-sensitive tasks like traffic rerouting, congestion pricing, or climate resilience planning~\cite{zhou2024dynamicroutegpt,guo2024explainable}.

\noindent \textbf{Challenges and Research Opportunities.}  
While modularized and RAG-enabled pipelines address several issues, challenges remain:
\begin{itemize}[leftmargin=*]
	\item \textbf{Spatial and Temporal Correlations:} Generative models must capture the intricate spatial-temporal dependencies inherent to transportation systems, such as dynamic OD flows, route interactions, and network bottlenecks. These correlations are further complicated by differences across urban environments, cultural driving habits, and localized traffic patterns. As much of transportation planning is inherently local, results produced by generative models often lack direct transferability. This necessitates careful localization, calibration, and validation to ensure that AI-driven insights align with the specific characteristics of the region being studied. For example, driving behaviors in Los Angeles, New York City, rural areas, or cities in Asia can vary significantly, underscoring the importance of adapting generative AI outputs to reflect these contextual nuances accurately. Without such adaptations, model predictions may fail to meet the practical requirements of local transportation planning.
	\item  \textbf{Domain Knowledge Integration:} Hybrid approaches that integrate generative AI with established transportation models, such as SUMO~\cite{lopez2018microscopic} and CityFlow~\cite{zhang2019cityflow} for microscopic traffic simulation or ActivitySim~\cite{gali2008activitysim} for activity-based travel demand modeling, are essential to ensure both realism and interpretability. While generative AI excels at synthesizing patterns from data, transportation planning often relies on domain knowledge, such as established traffic flow principles, behavioral models, and policy constraints. The integration of domain-specific models can guide generative AI systems to produce outputs that align with theoretical foundations and practical realities. For instance, combining AI-generated traffic scenarios with SUMO simulations ensures that outputs adhere to known traffic dynamics, while ActivitySim allows for a nuanced understanding of travel behavior at the individual and household levels. Such hybrid approaches reduce the risk of unrealistic outputs and improve the utility of generative AI for real-world decision-making.
	
	\item  \textbf{Real-Time Adaptability:} Scaling generative AI pipelines to handle rapidly evolving inputs, such as live sensor data, dynamic weather conditions, or incident reports, requires advancements in dynamic inference strategies. Transportation systems operate in real-time, with conditions changing by the second, necessitating models that can adapt quickly and continuously to new data streams. Current generative AI methods often struggle with this level of adaptability due to their reliance on static datasets or pre-trained models. Enhancing real-time adaptability involves developing techniques like online learning, which updates model parameters as new data arrives, or retrieval-augmented generation (RAG), which incorporates live data retrieval during inference. For example, dynamic inference could allow traffic management systems to instantly respond to accidents or congestion by recalibrating traffic signals and recommending alternate routes in real-time, ensuring system efficiency under continuously changing conditions.
	
	\item \textbf{Computational Scalability:} Scaling generative AI to support large-scale transportation networks requires resource-efficient methods, such as Low-Rank Adaptation (LoRA)~\cite{wu2024dlora} and agent-based distributed inference. Transportation systems are inherently complex, involving high-dimensional data across multiple regions, modes, and time periods. Traditional AI methods often face scalability issues when applied to such large-scale problems, resulting in prohibitive computational costs. Techniques like LoRA, which fine-tune only a subset of model parameters, significantly reduce computational overhead while maintaining performance. Additionally, agent-based distributed inference can parallelize computations across multiple nodes or devices, enabling real-time processing of expansive transportation networks. For example, distributed inference could optimize traffic signals across an entire metropolitan area by leveraging edge computing at individual intersections while maintaining centralized oversight for global system optimization. These scalable methods are critical for ensuring the practical deployment of generative AI in large-scale and resource-intensive transportation applications.  
\end{itemize}

Addressing these challenges will enable generative AI to deliver accurate, scalable, and context-aware solutions, transforming transportation systems into more responsive and sustainable networks.

\subsection{Data Scarcity and the Construction of Domain-Specific Datasets} \label{dataScarcity}

Data scarcity remains a critical barrier to deploying generative AI in transportation planning. Unlike computer vision or NLP, transportation research lacks large-scale, curated datasets tailored for tasks such as traffic flow prediction, OD calibration, and multimodal optimization~\cite{da2024open}. This scarcity arises from high data collection costs, privacy concerns, and the dynamic nature of transportation systems, which require continuous updates. Additionally, the fragmented nature of the transportation planning industry—divided by agency jurisdiction and geographic region—makes it difficult to create standardized, interoperable datasets on a nationwide scale. This lack of coordination limits the availability of shared datasets and hinders the scalability of generative AI applications. Future advancements in data-sharing agreements, standardized formats, and cross-agency collaboration will be essential to overcome these challenges and unlock the full potential of generative AI in transportation planning.

\noindent \textbf{High-Quality, Domain-Specific Dataset Development.}  
To address data scarcity, future work must focus on curating domain-specific datasets that reflect real-world transportation dynamics. Examples include:
\begin{itemize}[leftmargin=*]
	\item OD Calibration: Datasets such as OpenTI~\cite{da2024open} integrate OD matrices with observed traffic flows, supporting demand estimation tasks across different traffic environments. However, these datasets often require region-specific adaptations to accommodate local travel behaviors and network structures.
	\item Traffic Simulation: Simulation-based synthetic datasets using simulators~\cite{krajzewicz2012recent,zhang2019cityflow} can model vehicle interactions, congestion patterns, and route diversions under varying conditions. While useful for AI training, the realism of these datasets depends on their alignment with empirical traffic data from diverse locations.
	\item Multi-Modal Demand: Real-world datasets that integrate GPS logs, public transit feeds, and ride-sharing data provide valuable insights into multimodal mobility trends. However, existing datasets such as the NYC Taxi Dataset~\cite{nyctaxi2020} are often city-specific, limiting their generalizability. Expanding dataset coverage to diverse urban and rural contexts is necessary to ensure broader applicability.
\end{itemize}

\noindent \textbf{Synthetic Data Generation.}  
Generative AI provides a powerful solution for addressing data scarcity by creating synthetic datasets that replicate real-world transportation patterns. These datasets can fill critical gaps where historical data is unavailable, incomplete, or difficult to collect. For example, generative models can simulate OD matrices for rare events such as natural disasters or large-scale festivals, where traditional data collection methods fall short. Additionally, synthetic data can model mobility patterns in regions with limited transportation records, providing valuable insights for infrastructure planning and policy development.

To ensure the reliability of synthetic data, rigorous validation protocols are necessary. Metrics such as spatial consistency, trip distribution accuracy, and correlation with real-world datasets (e.g., OpenTI~\cite{da2024open}) can quantify the fidelity of generated data. Calibration techniques that align synthetic outputs with observed traffic trends further enhance their applicability in decision-making processes. Without proper validation, synthetic datasets risk introducing biases or unrealistic patterns that may compromise the effectiveness of transportation models.

\noindent \textbf{Collaborative Data Partnerships.}  
Public-private collaborations between transportation agencies, AI researchers, and urban planners are essential to overcoming data fragmentation and improving access to high-quality datasets. By standardizing data formats and implementing secure data-sharing agreements, collaborative platforms can facilitate the exchange of traffic, infrastructure, and mobility data while maintaining privacy compliance. These partnerships can also support the development of federated learning approaches, where models are trained across decentralized datasets without directly sharing sensitive information.

Efforts to improve interoperability between government databases, private mobility providers, and research institutions will be key to enabling scalable generative AI applications in transportation. By fostering a more integrated data ecosystem, these partnerships can ensure that AI-driven transportation models reflect diverse geographic contexts, evolving mobility patterns, and real-world infrastructure constraints.

\subsection{Enhancing Explainability and Reducing Hallucination Risks} \label{explain}
\noindent \textbf{Improving Explainability.}  
Ensuring the interpretability of AI-generated transportation insights is critical for decision-making. Generative AI models, particularly those used in OD calibration, travel demand forecasting, and policy simulations, must provide transparent reasoning behind their predictions. Techniques such as feature attribution and causal reasoning can significantly enhance explainability. Feature attribution methods, such as SHAP~\cite{antwarg2021explaining}, identify key inputs that influence AI-generated outputs, helping planners understand the relationship between traffic patterns and model predictions. Causal reasoning~\cite{chiunveiling,kiciman2023causal} allows AI models to go beyond correlation-based predictions by explicitly modeling cause-and-effect relationships, improving trust in AI-driven recommendations.

For example, in OD calibration, explainability techniques can highlight how congestion levels, travel demand, and socioeconomic factors contribute to the generated OD matrices, allowing planners to validate whether model outputs align with real-world mobility trends. Similarly, in policy recommendation systems, causal inference can provide insights into how interventions like congestion pricing or transit subsidies impact mode choice and travel behavior, enabling policymakers to evaluate potential trade-offs and unintended consequences before implementation.

\noindent \textbf{Reducing Hallucinations.}  
One of the major risks associated with generative AI is hallucination—the generation of outputs that appear plausible but lack factual grounding~\cite{perkovic2024hallucinations,yao2023llm}. In transportation, hallucinations can manifest in erroneous traffic predictions, unrealistic infrastructure recommendations, or misleading policy analyses, all of which can lead to costly misallocations of resources. Addressing hallucination risks requires a combination of verification mechanisms, dynamic retrieval methods, and uncertainty quantification.

Validation Checkpoints serve as critical safeguards to cross-verify AI-generated outputs against ground-truth data sources. For example, in traffic forecasting, models can be cross-validated against sensor data and historical congestion patterns to prevent spurious predictions~\cite{dhuliawala2023chain}. In policy simulations, AI outputs should be compared with real-world case studies to assess feasibility and relevance.

Agentic Retrieval-Augmented Generation (RAG) techniques enhance AI reliability by dynamically incorporating real-time data during inference. Instead of relying solely on pre-trained knowledge, RAG retrieves live updates from databases, government reports, or transit feeds to ground AI-generated responses in factual knowledge~\cite{bechard2024reducing,niu2023ragtruth, da2024evidencechat, dai2024vistarag}. For instance, in real-time travel demand modeling, RAG-enabled AI can retrieve up-to-date ridership statistics or ride-hailing activity, ensuring predictions reflect current mobility trends rather than outdated assumptions.

Uncertainty Quantification techniques, such as probabilistic variance estimation and confidence calibration, can further enhance AI trustworthiness by flagging predictions with high uncertainty~\cite{lin2023generating,da2024llm,varshney2023stitch}. In network optimization tasks, generative AI can assess the confidence of alternative route recommendations, allowing planners to prioritize high-certainty solutions while further scrutinizing uncertain outputs. Similarly, uncertainty-aware models in traffic signal optimization can indicate whether AI-generated timing adjustments are robust or require additional validation from human experts.

By integrating these strategies, generative AI can provide more interpretable, trustworthy, and actionable transportation insights while minimizing risks associated with hallucinations and unreliable outputs. These improvements are essential to ensuring that AI-driven transportation models remain credible, robust, and aligned with real-world operational constraints.

\subsection{Democratizing Access to Transportation Knowledge}\label{democratizing}

Generative AI has the potential to bridge the gap between technical transportation expertise and public accessibility by simplifying complex data, enabling multilingual support, and fostering citizen engagement. Traditional transportation planning reports and policy documents are often highly technical, limiting public participation and stakeholder involvement. AI-powered tools can generate easy-to-understand summaries of infrastructure projects, congestion management plans, and policy changes, making them more accessible to non-expert audiences.

Multilingual AI systems further enhance inclusivity by translating critical transportation information, such as traffic updates, road closures, and urban development plans, into multiple languages. This capability ensures that non-English-speaking communities can stay informed about mobility decisions that affect their daily lives.

Beyond passive information dissemination, generative AI can facilitate two-way engagement through interactive citizen platforms. AI-driven public forums or participatory urban planning interfaces can synthesize feedback from diverse stakeholders, allowing transportation agencies to gauge public sentiment on transit projects, pedestrian-friendly initiatives, or multimodal integration strategies. Such platforms ensure that mobility solutions align with community needs and promote equity in transportation decision-making.

By reducing technical barriers and enabling real-time public engagement, generative AI fosters a more inclusive and participatory approach to transportation planning, ensuring that mobility solutions reflect the diverse needs of urban and rural communities alike.

\subsection{Call for Novel Evaluation Criteria for Generative AI}\label{evaluation}

Existing evaluation metrics such as RMSE, BLEU~\cite{papineni-etal-2002-bleu}, and standard accuracy benchmarks are insufficient for assessing the real-world applicability of generative AI in transportation systems~\cite{srinivasan2023beyond}. Unlike traditional NLP or computer vision tasks, transportation applications require a broader evaluation framework that considers system-level impacts, fairness, and interpretability.

A comprehensive set of metrics must be established to ensure AI-generated outputs align with transportation planning goals. System efficiency should be measured through congestion reduction, travel time savings, and fuel consumption improvements, ensuring that AI-driven recommendations optimize mobility networks. Sustainability impact must be incorporated by assessing reductions in carbon emissions and modal shifts toward environmentally friendly transportation options. Equity metrics are critical to evaluating whether AI-generated policies and infrastructure improvements benefit underserved communities and mitigate disparities in transportation accessibility. Additionally, interpretability and uncertainty quantification should be prioritized to ensure AI-generated insights are transparent, trustworthy, and actionable for planners and policymakers.

By expanding evaluation criteria beyond traditional performance benchmarks, researchers can better align generative AI advancements with the needs of transportation planning, ensuring that AI-driven solutions are efficient, equitable, and adaptable to evolving urban mobility challenges. Addressing these challenges through modular pipelines, robust data curation, and explainability frameworks will unlock generative AI's full potential in building sustainable and intelligent transportation systems.

\section{Conclusion}
\label{sec:conclusion}

This survey provides a comprehensive interdisciplinary exploration of the integration of generative AI into transportation planning, bridging the gap between classical methodologies and modern computational approaches. We introduce a novel principled taxonomy that systematically categorizes transportation tasks and generative AI-driven methods, offering a structured framework to guide researchers and practitioners in leveraging AI effectively. 
We begin by detailing generative AI capabilities across predictive modeling, synthetic data generation, simulation of traffic dynamics, and societal impacts, emphasizing their transformative potential for real-world transportation planning and operations. Our analysis highlights both opportunities and challenges, including data scarcity, biases, explainability limitations, and ethical concerns in deploying generative AI systems.
Furthermore, we explore computational techniques tailored to transportation planning, including domain-specific fine-tuning strategies to enhance performance on tasks like \textit{origin-destination (OD) calibration} and \textit{traffic signal optimization}, advanced inference methods such as \textit{zero-shot} and \textit{few-shot learning} for dynamic traffic predictions and agent-based simulations, the integration of retrieval-augmented generation (RAG) and modular pipelines to incorporate real-time transportation data for actionable insights, and the creation of benchmark datasets and synthetic traffic analysis, ensuring robust evaluation frameworks for generative AI applications. An empirical case study on OD flow calibration demonstrates the practical implementation of these methods, illustrating the value of generative AI in improving the efficiency and scalability of complex transportation systems.

Finally, we identify critical research directions to address the open challenges in the field:
\begin{itemize}[leftmargin=*]
	\item There is a need for modular pipelines that integrate domain knowledge with generative AI methods in a way that minimizes the technical burden on practitioners, ideally through GenAI-as-a-service solutions that do not require deep expertise in AI.
	\item Future work should focus on developing evaluation metrics that are directly aligned with transportation system goals, such as \textit{sustainability}, \textit{system efficiency}, and \textit{mobility equity}.
	\item robust methodologies are required to handle \textit{uncertainty quantification}, \textit{bias mitigation}, and the adoption of \textit{ethical AI frameworks}.
	\item Democratizing access to transportation planning and analysis via AI-powered tools, such as multilingual interfaces and simplified model explanations, can empower diverse stakeholder groups and enhance participatory decision-making..
\end{itemize}

\vspace{1mm}
\noindent \textbf{Final Outlook.} Generative AI is poised to reshape transportation planning by offering data-driven solutions for predictive modeling, simulation, and real-time operations. Realizing its full potential will require addressing key challenges such as system alignment, ethical use, and practical accessibility. Interdisciplinary collaboration between AI researchers, transportation engineers, and policymakers will be essential to ensure that generative tools are both effective and equitable. Looking ahead, we envision a future where generative AI serves as a core enabler of adaptive, efficient, and sustainable mobility systems, especially in under-resourced or rapidly evolving urban environments. We hope this survey can serve as a starting point for mutual understanding between AI and transportation planning domain for continued research and deployment of generative AI approaches tailored for transportation challenges, with the goal of enhancing mobility, equity, and quality of life at scale.

\bibliographystyle{unsrt}
\bibliography{references}

\clearpage
\appendix

\end{document}